\documentclass[acmtog,nonacm,screen]{acmart}
\renewcommand\footnotetextcopyrightpermission[1]{} %
\AtBeginDocument{%
  \providecommand\BibTeX{{%
    \normalfont B\kern-0.5em{\scshape i\kern-0.25em b}\kern-0.8em\TeX}}}

\usepackage{graphicx}
\usepackage[capitalize]{cleveref}

\usepackage{orcidlink}

\usepackage{wrapfig}
\usepackage{multirow}
\usepackage{tabularx}
\usepackage{algorithmic}
\usepackage[ruled]{algorithm2e}
\usepackage{mathtools}
\usepackage{balance}

\usepackage{float}
\usepackage{enumitem}
\usepackage{comment}
\usepackage{pifont}
\usepackage[toc,page]{appendix}

\usepackage{soul}
\usepackage{acronym}
\usepackage{fancybox}
\usepackage{epigraph}
\usepackage{dirtytalk}
\usepackage{bm}
\usepackage{fontawesome}
\usepackage{colortbl}
\usepackage{booktabs}
\usepackage{xspace}

\usepackage{xr}
\usepackage{tikz}
\usepackage{microtype}
\usepackage{transparent}
\usepackage{subfig}
\usepackage{multicol}
\usepackage[htt]{hyphenat}
\usepackage[toc,page]{appendix}

\definecolor{citecolor}{HTML}{0071bc}
\definecolor{frontcolor}{HTML}{325ea5}
\definecolor{backcolor}{HTML}{a58b77}
\definecolor{sidecolor}{HTML}{10768c}
\definecolor{skincolor}{HTML}{dcb7b7}
\definecolor{darkred}{rgb}{0.6, 0.1, 0.05}
\definecolor{DeltaColor}{rgb}{0.039,0.73,0.71}
\definecolor{SigmaColor}{rgb}{0.98,0.45,0.0}
\definecolor{AlphaColor}{rgb}{0,0,0.8}
\definecolor{BetaColor}{rgb}{0.8,0,0.8}
\definecolor{GammaColor}{rgb}{0.514,0.34,0.224}
\definecolor{EpsilonColor}{rgb}{0.353,0.725,0.906}
\definecolor{PurpleColor}{HTML}{8B008B}
\definecolor{BadColor}{HTML}{C0392B}
\definecolor{OrangeColor}{rgb}{0.914,0.541,0.0.141}
\definecolor{GreenColor}{HTML}{00ab41}
\definecolor{RedColor}{rgb}{0.949,0.275, 0.224}
\definecolor{LightCyan}{rgb}{0.88,1,1}
\definecolor{Gray}{gray}{0.85}
\definecolor{LightGray}{gray}{0.70}

\definecolor{greenprior}{HTML}{34a853}
\definecolor{redprior}{HTML}{ea4335}
\definecolor{blueprior}{HTML}{4285f4}

\definecolor{bestcolor}{rgb}{1, 0.5, 0.25}
\definecolor{secondbestcolor}{rgb}{1, 0.8, 0.5}

\newcommand{\etc}{\mbox{etc}\xspace}
\newcommand{\etal}{\mbox{et al.}\xspace}
\newcommand{\ie}{\mbox{i.e.}\xspace}
\newcommand{\eg}{\mbox{e.g.}\xspace}

\newcolumntype{a}{>{\columncolor{Gray}}c}

\newcommand{\qheading}[1]{\noindent\textbf{#1.}}

\newcommand{\zheading}[1]{\textbf{#1.}}

\makeatletter
\newcommand*{\addFileDependency}[1]{%
  \typeout{(#1)}
  \@addtofilelist{#1}
  \IfFileExists{#1}{}{\typeout{No file #1.}}
}
\makeatother

\newlength\savewidth\newcommand\shline{\noalign{\global\savewidth\arrayrulewidth
  \global\arrayrulewidth 1pt}\hline\noalign{\global\arrayrulewidth\savewidth}}

\newcommand{\vid}{\href{https://www.youtube.com/watch?v=sylXTxG_U2U}{\textcolor{magenta}{video}}\xspace}
\newcommand{\supmat}{SupMat.\xspace}

\newcommand{\specific}[1]{\xspace{\textit{\small{\fontfamily{qcr}\selectfont{#1}}}}\xspace}

\newcommand{\geowizard}{\mbox{GeoWizard}\xspace}
\newcommand{\genpercept}{\mbox{GenPercept}\xspace}

\newcommand{\sota}{state-of-the-art\xspace}

\newcommand{\modelname}{\textcolor{black}{StableNormal}\xspace}
\newcommand{\modelnameLong}{Reducing Diffusion Variance for Stable and Sharp Normal}
\newcommand{\ourtitle}{\modelname: \modelnameLong}

\newcommand{\benchmarkURL}{\href{https://huggingface.co/spaces/Stable-X/normal-estimation-comparison}{\specific{\textcolor{magenta}{Compare}}}}
\newcommand{\demoURL}{\href{https://huggingface.co/spaces/Stable-X/StableNormal}{\specific{\textcolor{magenta}{Demo}}}}

\acrodef{amt}[AMT]{Amazon Mechanical Turk}

\begin{document}

\title{\ourtitle}

\author{Chongjie Ye}
\authornote{Equal Contribution}
\email{chongjieye@link.cuhk.edu.cn}
\orcid{0000-0002-7123-0220}
\author{Lingteng Qiu}
\authornotemark[1]
\email{220019047@link.cuhk.edu.cn}
\orcid{0000-0002-3250-0486}
\affiliation{%
  \institution{The Chinese University of Hongkong, Shenzhen}
  \country{China}
}

\author{Xiaodong Gu}
\email{dadong.gxd@alibaba-inc.com}
\orcid{0000-0003-2623-7973}
\affiliation{%
  \institution{Alibaba Group}
  \country{China}
}
\author{Qi Zuo}
\email{muyuan.zq@alibaba-inc.com}
\orcid{0009-0006-2711-9767}
\affiliation{%
  \institution{Alibaba Group}
  \country{China}
}
\author{Yushuang Wu}
\email{yushuangwu@link.cuhk.edu.cn}
\orcid{0009-0002-9725-0606}
\affiliation{%
  \institution{The Chinese University of Hongkong, Shenzhen}
  \country{China}
}
\author{Zilong Dong}
\email{list.dzl@alibaba-inc.com}
\orcid{0000-0002-6833-9102}
\affiliation{%
  \institution{Alibaba Group}
  \country{China}
}
\author{Liefeng Bo}
\email{liefeng.bo@alibaba-inc.com}
\orcid{https://scholar.google.com/citations?user=FJwtMf0AAAAJ&hl=en}
\affiliation{%
  \institution{Alibaba Group}
  \country{China}
}
\author{Yuliang Xiu}
\authornote{Corresponding Author}
\email{yuliang.xiu@tuebingen.mpg.de}
\orcid{0000-0003-0165-5909}
\affiliation{%
  \institution{Max Planck Institute for Intelligent Systems}
  \country{Germany}
}
\author{Xiaoguang Han}
\authornotemark[2]
\email{hanxiaoguang@cuhk.edu.cn}
\orcid{0000-0003-0162-3296}
\affiliation{%
  \institution{The Chinese University of Hongkong, Shenzhen}
  \country{China}
}

\renewcommand{\shortauthors}{Ye, \etal}

\begin{abstract}

This work addresses the challenge of high-quality surface normal estimation from monocular colored inputs (\ie, images and videos), a field which has recently been revolutionized by repurposing diffusion priors. However, previous attempts still struggle with stochastic inference, conflicting with the deterministic nature of the Image2Normal task, and costly ensembling step, which slows down the estimation process. 
Our method, \modelname, mitigates the stochasticity of the diffusion process by reducing inference variance, thus producing ``Stable-and-Sharp'' normal estimates without any additional ensembling process. \modelname works robustly under challenging imaging conditions, such as extreme lighting, blurring, and low quality. It is also robust against transparent and reflective surfaces, as well as cluttered scenes with numerous objects.
Specifically, \modelname employs a coarse-to-fine strategy, which starts with a one-step normal estimator (YOSO) to derive an initial normal guess, that is relatively coarse but reliable, then followed by a semantic-guided refinement process (SG-DRN) that refines the normals to recover geometric details. 
The effectiveness of \modelname is demonstrated through competitive performance in standard datasets such as DIODE-indoor, iBims, ScannetV2 and NYUv2, and also in various downstream tasks, such as surface reconstruction and normal enhancement. These results evidence that \modelname retains both the \textit{``stability''} and \textit{``sharpness''} for accurate normal estimation. \modelname represents a baby attempt to repurpose diffusion priors for \textit{deterministic estimation}. To democratize this, code and models have been publicly available in \href{https://huggingface.co/Stable-X}{\specific{\textcolor{magenta}{hf.co/Stable-X}}}.

\end{abstract}


\begin{teaserfigure}
  \includegraphics[width=\linewidth]{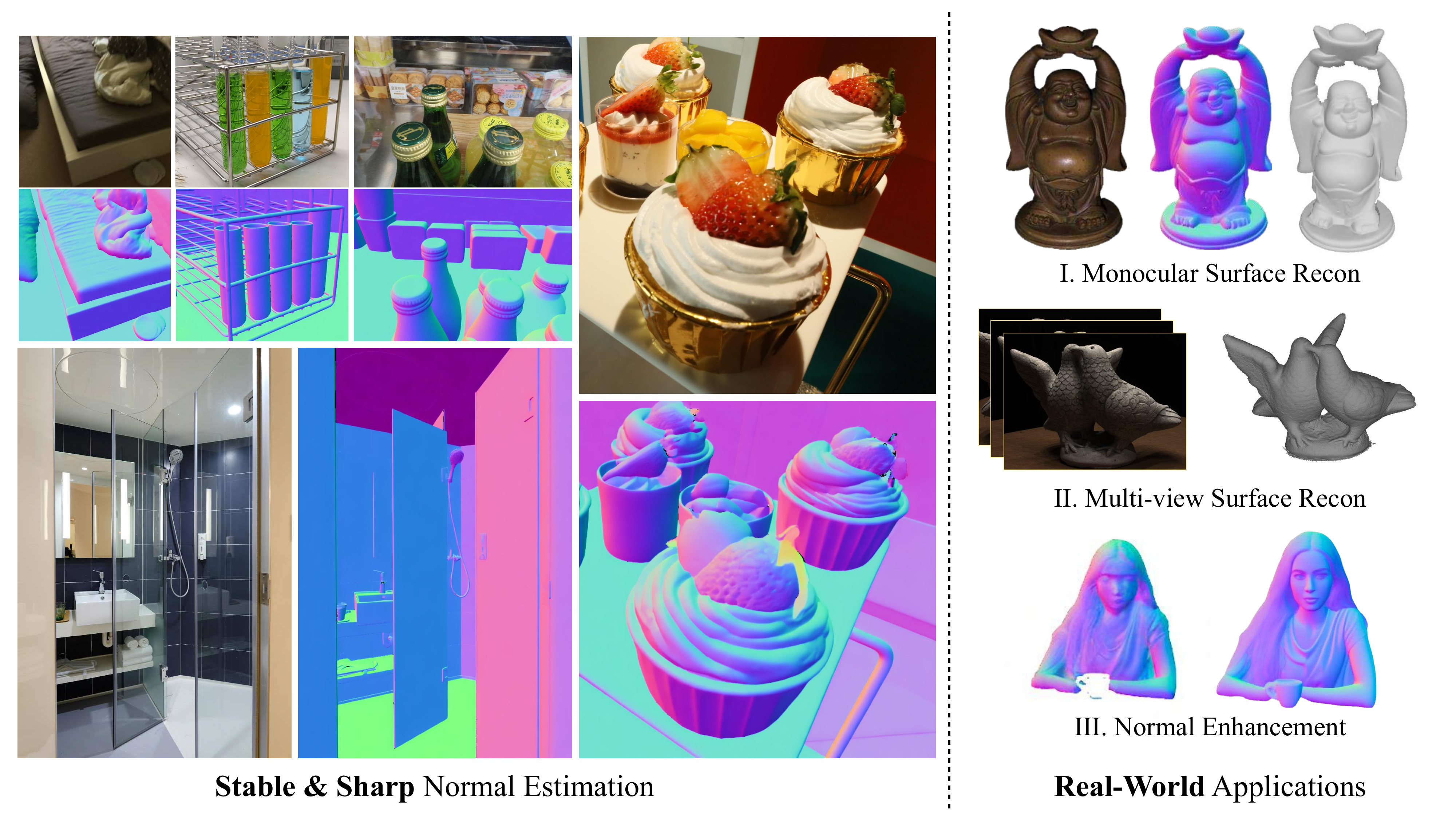}
  \caption{We propose \textbf{\modelname}, which tailors the diffusion priors for monocular normal estimation. Unlike prior diffusion-based works, we focus on enhancing estimation stability by reducing the inherent stochasticity of diffusion models (\ie, Stable Diffusion~\cite{rombach2021highresolution}). This enables \textit{``Stable-and-Sharp''} normal estimation, which outperforms multiple baselines (try~\benchmarkURL), and improves various real-world applications (try~\demoURL).}
  \label{fig:teaser}
\end{teaserfigure}

\maketitle

\newcommand{\Tref}[1]{Table~\ref{#1}}
\newcommand{\tref}[1]{Tab.~\ref{#1}}
\newcommand{\eref}[1]{Eq.~\eqref{#1}}
\newcommand{\Eref}[1]{Equation~\eqref{#1}}
\newcommand{\fref}[1]{Fig.~\ref{#1}}
\newcommand{\Fref}[1]{Figure~\ref{#1}}
\newcommand{\sref}[1]{Sec.~\ref{#1}}
\newcommand{\Sref}[1]{Section~\ref{#1}}
\newcommand{\DecoupleLoss}{Shrinkage Regularizer\xspace}
\newcommand{\DINOGuiderAbs}{SG-DRN\xspace}
\newcommand{\oneshot}{YOSO\xspace}
\newcommand{\oneshotName}{You-Only-Sample-Once\xspace}
\newcommand{\DINOGuider}{Semantic-Guided Diffusion Refinement Network\xspace}
\newcommand{\UNET}{${\mu}_\theta$\xspace}
\newcommand{\DUNET}{$\mu_\zeta$\xspace}
\newcommand{\bfparagraph}[1]{\textbf{#1}}

\newcommand{\noise}{$\boldsymbol{\epsilon}$}

\newcommand{\best}{\cellcolor{orange}}
\newcommand{\sbest}{\cellcolor{orange!50}}
\newcommand{\tbest}{\cellcolor{orange!20}}

\newcommand{\tdown}{$\downarrow$}
\newcommand{\tup}{$\uparrow$}
\newcommand{\marigold}{Marigold}

\newcommand{\uneteps}{${\boldsymbol{\mu}_{\theta}}^{\boldsymbol{\epsilon}}$\xspace}
\newcommand{\initT}{$x_{t^+}$\xspace}

\section{Introduction}

Normal map, as a 2.5D representation, bridges 2D and 3D worlds.
In 3D modeling, object surfaces are typically represented by polygons. Normal maps add illusory surface details to these polygons, which enhances their realism. In 2D domain, if accurately estimated from in-the-wild pixels, tasks such as relighting or intrinsic decomposition become feasible, opening the door to a broad spectrum of applications. \modelname aims to estimate \textit{accurate \& sharp} surface normals from monocular colored inputs (\ie, images, videos). 

In the era of deep learning, this ``Image2Normal'' task has been well explored in a line of works~\cite{fouhey2013data,eftekhar2021omnidata,eigen2015predicting,bansal2016marr,ranftl2021vision,wang2015designing}. Recently, advances in diffusion-based image generator, often trained on large-scale datasets~\cite{schuhmann2022laion}, have shifted the vision community's focus towards repurposing the diffusion priors~\cite{rombach2022stable} to estimate the geometric or intrinsic cues, such as depth~\cite{ke2023marigold}, normal~\cite{fu2024geowizard}, and materials~\cite{kocsis2024intrinsic}. 

\begin{figure}[t]
  \includegraphics[width=0.8\linewidth]{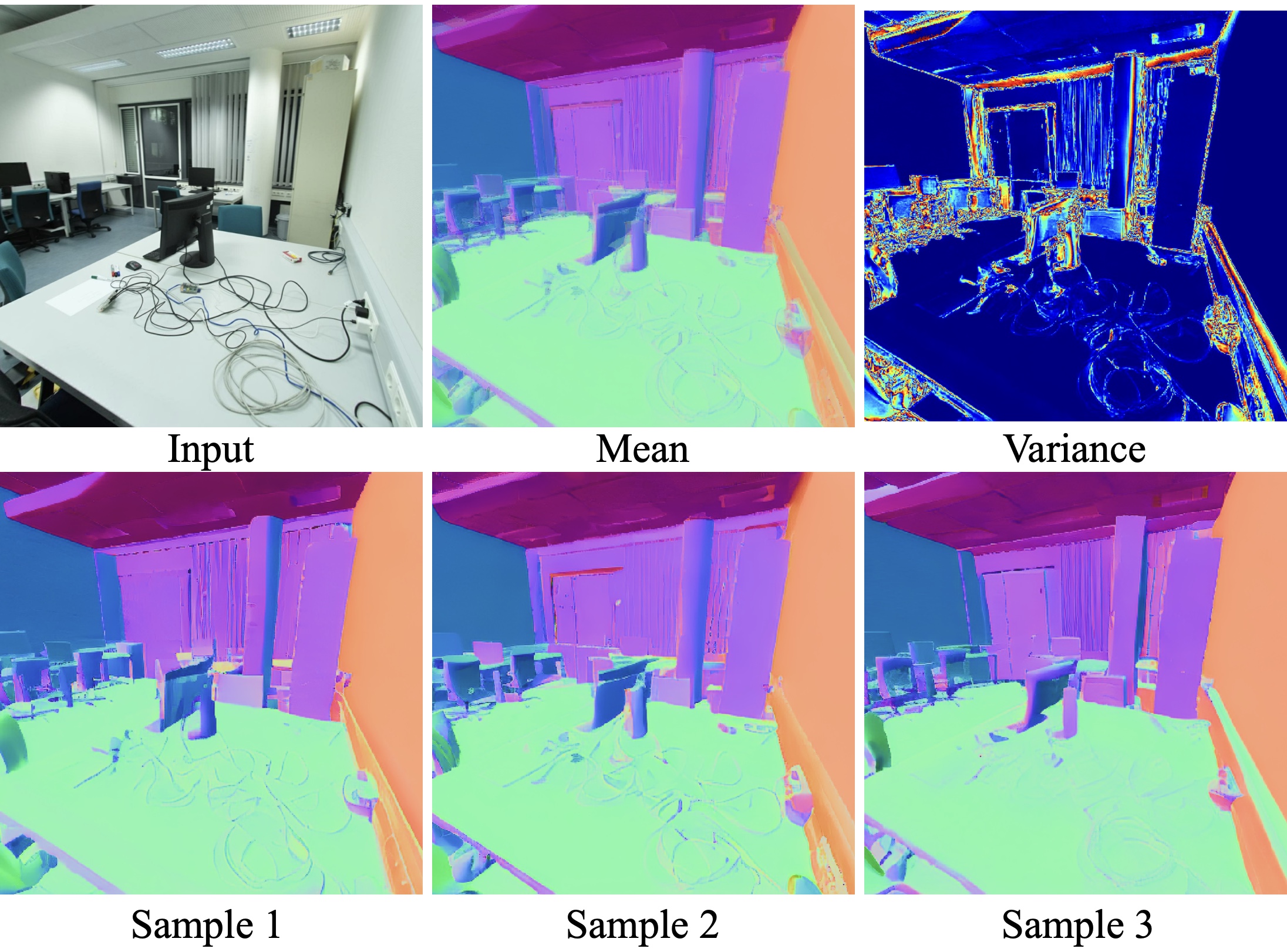}
  \caption{\textbf{High-variance normal estimations.} We show multiple samples for a single scene and visualize the mean and variance of the predicted normals. For each sample, while the normal maps exhibit sharp details, there is high variance in areas with high-frequency content. This high variance in sharp regions makes the inference less reliable.}
  \label{fig:fake-sharp}
\end{figure}

These efforts have yielded ``sharp-looking'' results (\cref{fig:fake-sharp}). However, human eyes lack the sensitivity to \textit{accurately} perceive the normal maps. Despite producing ``sharp-looking'' normals, temporal inconsistency exists~\footnote{\url{huggingface.co/docs/diffusers/main/en/using-diffusers/marigold_usage\#frame-by-frame-video-processing-with-temporal-consistency}}, and the results, even after being ensembled, still deviate significantly from ground-truth normals (\cref{fig:fake-sharp}). Simply put, these results are \textit{``sharp''} but neither \textit{``correct''} nor \textit{``stable''}. 

We attribute this to two factors: 1) unstable imaging conditions, such as extreme lighting, dramatic camera movement, motion blur, and low-quality images. 2) inductive bias of the diffusion process --- stochasticity. Such stochasticity contradicts the nature of the estimation process, which should be as deterministic as possible.
Therefore, a crucial question is raised: 

\medskip
\textit{How can we mitigate the inherent \underline{stochasticity} of the diffusion process for \underline{deterministic} estimation?}
\medskip

\begin{figure}[t]
  \includegraphics[trim=000mm 020mm 000mm 000mm, clip=true, width=\linewidth]{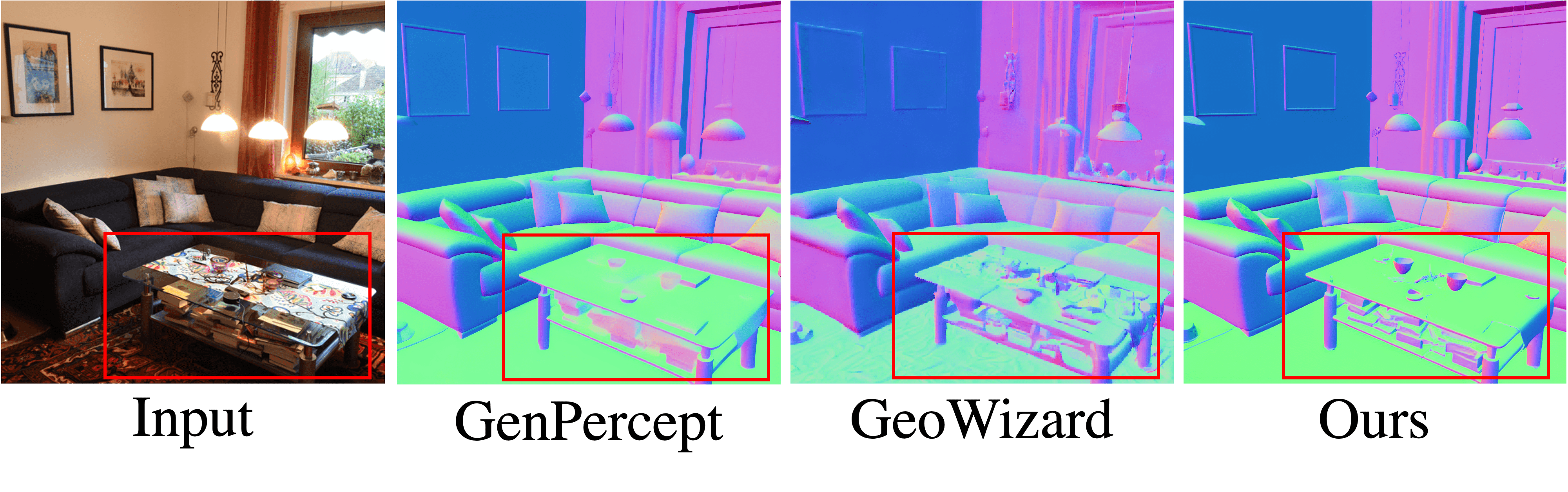}
  \small
  \begin{tabularx}{\linewidth}{
 >{\centering\arraybackslash}X
 >{\centering\arraybackslash}X
 >{\centering\arraybackslash}X
 >{\centering\arraybackslash}X
}
Input & \genpercept & \geowizard & Ours\\
 & \cite{xu2024diffusion} & \cite{fu2024geowizard} & \\

 \end{tabularx}
  \caption{\textbf{Comparative Analysis of Normal Estimators: ``Stability'' vs. ``Sharpness''.} One-step \textit{\genpercept} compromises the high-frequency details and produces overly-smooth normals for objects on the table, while \textit{\geowizard} produces seemingly sharp normals, but neither correct nor stable. Our method well balances stability and sharpness. The \fcolorbox{red}{white}{red boxes} highlight the visual difference mentioned above.}
  \label{fig:stable-shaprness}
\end{figure}

Answering this question in the normal domain is more urgent than in depth domain. Since monocular depth estimation typically estimates affine-invariant depth (\ie, depth values up to a global offset and scale), while surface normals are not subject to scale and translation ambiguity. That is to say, given a single image, the task of normal estimation (one-to-one mapping), is more ``deterministic'' than depth estimation (one-to-many mapping). However, eliminating stochasticity from the diffusion process, like the one-step \genpercept~\cite{xu2024diffusion}, could compromise the recovery of high-frequency details and result in overly-smooth normals~(See~\cref{fig:stable-shaprness}). 
Thus, finding a balance between \textit{``stability''} vs. \textit{``sharpness''} is needed.

So we present \modelname to tackle this trade-off. It demonstrates that, a reliable initialization, coupled with a stable refinement, is essential to produce sharp and stable normal estimates. Our approach follows the coarse-to-fine scheme: 1) one-step normal estimation (\cref{sec:yoso}) for reliable initialization, and 2) semantic-guided diffusion refinement (\cref{sec:sg-refine}) to progressively sharpen the normal maps in a semantic-aware direction.

Specifically, a \DecoupleLoss is introduced to train the one-step normal estimator, which reduces the training variance by splitting the vanilla diffusion loss into generative and reconstruction terms. This one-step estimator, namely YOSO (You-Only-Sample-Once), already performs on-par with current \sota DSINE~\cite{bae2024dsine}, see~\cref{table:ablation}. 
Additionally, \DINOGuider~(\DINOGuiderAbs) is presented to enhance the stability of the diffusion-based refinement process by integrating DINO semantic priors. Such priors decrease sampling variance while enhancing local details, as shown in \cref{fig:ablations_DINO}.

We evaluate \modelname on DIODE-indoor, iBims, ScannetV2, and NYUv2 datasets. Also, we show how our strong normal estimator improves various reconstruction scenarios (\ie, object-level, indoor-scene, and normal integration-based). The superiority of \modelname~is substantiated both qualitatively and quantitatively. Please check the \vid and~\cref{fig:teaser} to see how robust \modelname performs in challenging conditions, such as extreme lighting, blurring, object transparency \& reflections, or clustered scenes.

\medskip
The main contributions of \modelname are as follows:

\begin{itemize}
    \item We pinpoint the \textbf{critical issue} why diffusion priors cannot be directly (w/o bells and whistles, \eg post-ensembling) applied on ``Image2Normal'' task --- the inherent conflict between the \textit{``stochastic''} diffusion process and \textit{``deterministic''} requirement for geometric cues estimation.
    \item To address this conflict, we propose a \textbf{simple-yet-effective solution}, namely ``\modelname''. It justifies that a reliable initialization (YOSO), coupled with a stable refinement (\DINOGuiderAbs), is essential to estimate sharp normals steadily.
    \item We conduct \textbf{extensive experiments} to evaluate \modelname's accuracy. It not only outperforms other baselines by a large margin in high-quality indoor benchmarks (\ie, DIODE-indoor, iBims, and ScannetV2), but also far ahead of its peers (\ie, \geowizard, DSINE) in terms of inference stability at real-world scenarios, even under extreme conditions. This stability benefits many downstream tasks, see~\cref{fig:teaser}.
\end{itemize}

\section{Related Works}
\subsection{Regression-based Monocular Normal Estimation}
Surface normal estimation from monocular RGB inputs has been extensively studied ~\cite{Fouhey_Gupta_Hebert_2013, Fouhey_Gupta_Hebert_2014, Ladicky_Zeisl_Pollefeys_2014, Wang_Fouhey_Gupta_2015, Eigen_Fergus_2015, Wang_Shen_Russell_Cohen_Price_Yuille_2016, Qi_Liao_Liu_Urtasun_Jia_2018, Huang_Zhou_Funkhouser_Guibas_2019, Zhang_Cui_Xu_Yan_Sebe_Yang_2019, Liao_Gavves_Snoek_2019, Qi_Liu_Liao_Torr_Urtasun_Jia_2022, Do_Vuong_Roumeliotis_Park_2020, Wang_Geraghty_Matzen_Szeliski_Frahm_2020}. In general, the prior regression-based methods consist of a feature extractor, followed by a prediction head. Hoiem \etal~\cite{Hoiem_Efros_Hebert_2005, Hoiem_Efros_Hebert_2007} were the pioneers in framing this classic task as a statistical learning problem. The output space was discretized, and handcrafted features were extracted to classify the normals. However, such features are generally designed for specific scenarios and cannot generalize well to unseen scenes. 

This generalization problem was later addressed by deep learning techniques in a data-driven manner~\cite{Bansal_Russell_Gupta_2016, Wang_Fouhey_Gupta_2015}. More recently, Omnidata-V2~\cite{eftekhar2021omnidata}, with a U-Net architecture~\cite{Ronneberger_Fischer_Brox_2015}, is trained on a large-scale data (12M) captured from diverse scenes under various camera settings. Bae~\etal~\cite{Bae_Budvytis_Cipolla_2021} propose to estimate the per-pixel surface normal probability distribution, from which the expected angular error can be inferred to quantify the aleatoric uncertainty. The transition from CNNs to vision transformers (ViT) has further advanced this field, as demonstrated by DPT~\cite{Ranftl_Bochkovskiy_Koltun_2021}. DSINE~\cite{bae2024dsine} rethinks how to correctly model the inductive biases for surface normal estimation, and proposes to leverage the per-pixel ray direction, and learn the relative rotation between nearby pixels. These efforts decrease the need for large-scale training data, DSINE trained only on 160K images surpass the Omnidata-V2, which is trained on over 12M images. Recently, inspired by visual prompting~\cite{bar2022visual}, background prompting~\cite{baradad2023background} was introduced to reduce the domain gap between synthetic and real data, by simply placing the segmented object into a learned background ``prompt''. 
Despite steady advancements, regression-based normal estimators, trained on limited and constrained data, continue to face generalization issues and struggle to capture fine-grained geometric details.
 
\subsection{Diffusion-based Monocular Normal Estimation}
Recently, the computer vision community has witnessed the bloom of diffusion-based Text-to-Image (T2I) model and its extensions~\cite{rombach2021highresolution, zhang2023adding, Peebles_Xie_2022}. Several works have explored how to adapt the strong pretrained model, thus repurpose it as geometric cues estimator~\cite{ji2023ddp,ke2023repurposing,zhao2023unleashing,liu2023hyperhuman,long2023wonder3d,qiu2023richdreamer,fu2024geowizard}. Wonder3D~\cite{long2023wonder3d} proposes to model the joint distribution of color and normal to enhance their consistency, which has been shown to improve the quality of the final 3D output. Richdreamer~\cite{qiu2023richdreamer} concurrently trains a depth and normal diffusion model on the large-scale LAION-2B dataset~\cite{schuhmann2022laion}, utilizing predictions from the off-the-shelf normal and depth estimators~\cite{lasinger2019towards}. Moreover, Geowizard~\cite{fu2024geowizard} extends Wonder3D by adding a geometry switcher (indoor/outdoor/object) to segregate the multi-sourced data distribution of various scenes into distinct sub-distributions.

Although these diffusion-based approaches can capture \textit{``sharp-looking''} surface details, these results actually deviates significantly from ground-truth in normal space, owing to to the inherent high-variance of diffusion process~(see~\cref{fig:fake-sharp}). The large variance is first introduced by Gaussian initialization, which is propagated and amplified in the entire multi-step diffusion process (\ie, signal-leak issue~\cite{everaert2024exploiting}). 
In fact, some prior research has explored this issue, either employing an affine-invariant ensembling strategy during the post-processing stage~\cite{ke2023marigold,fu2024geowizard}, or completely discarding the iterative multi-step generation process, thus shifting towards a one-step perception problem~\cite{xu2024diffusion}. 

However, both strategies come with their own pitfalls: post-ensembling, which applies to multiple outputs, is computationally intensive. The assumption of affine invariance often fails to generalize across different types of outputs, like normals.
While GeoWizard~\cite{fu2024geowizard} exhibits sharper results compared to other traditional approaches, it does not notably improve quantitative performance, suggesting that diffusion-based normal estimators induce the directional deviation in normal space (see~\cref{fig:fake-sharp}).  
Furthermore, without the post-ensembling step, the diffusion-based estimators tend to produce outputs with large variance (see~\cref{fig:fake-sharp}), highlighting its \textbf{inherent stochastic nature}.
Regarding the one-step approach, it oversimplifies the markov chain of the diffusion process, smoothing out intrinsic local geometric details, leading to the typical \textbf{over-smoothing artifacts} seen in other regression-based methods~\cite{bae2024dsine,eftekhar2021omnidata}. 
Therefore, when repurposing the diffusion model for deterministic estimation tasks, such as normal estimation, a trade-off between \textit{``stability''} and \textit{``sharpness''} arises, which requires careful consideration before proceeding.

\section{Method}
\label{sec:method}

\begin{figure*}[t] \centering
    \includegraphics[width=\textwidth]{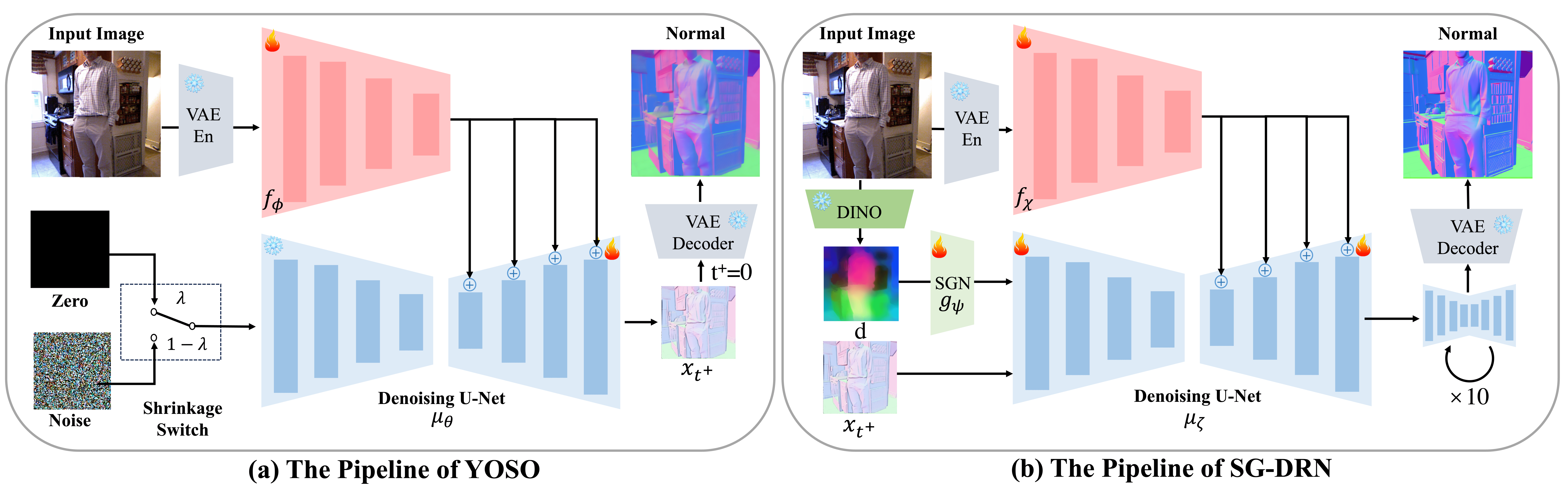}
    \caption{
        \textbf{Overview of the \modelname}. 
     The overall pipeline is composed of two stages: 1) \oneshot aims to produce a confident initialization \initT for stage two with a novel \DecoupleLoss; 2) \DINOGuiderAbs plays the role of stable denoising, by leveraging the stronger semantic control information extracted from DINO~\cite{oquab2024dinov2}. The textual prompt for the U-Net in both stages is set to ``\specific{The normal map}''. } 
    \label{fig:pipeline}
\end{figure*}

\subsection{Preliminaries on Diffusion Model}

Diffusion Probabilistic Models~\cite{ho2020denoising, song2020denoising} aim to model a data distribution $p(x)$ by sequentially transform a Gaussian distribution via the so-called backward diffusion process $x_{t-1} = B_t x_t - \mu_\theta^{\epsilon}(x_t, t) + \epsilon_t$ in which $\epsilon_t \sim \mathcal{N}(0, \sigma_tI)$ and $\mu_\theta^{\epsilon}$ predicts the injected noise. This backward process is uniquely determined by a predefined forward diffusion process $x_{t+1} = A_t x_t + \epsilon_t$.

As a classical example, DDPM~\cite{ho2020denoising} assumes that the initial Gaussian distribution $\mathcal{N}(\mathbf{0}, \mathbf{I})$ can be obtained by running the following forward diffusion process:

\begin{equation}
q(x_t) = \sqrt{\alpha_t} {x_0} + \sqrt{1-\alpha_t} \boldsymbol{\epsilon}, x_0\sim p(x), t\in \{0, 1, ..., T\}
\label{eq:q_sampling}
\end{equation} 
where $\bm{\epsilon} \sim \mathcal{N}(\mathbf{0}, \mathbf{I})$, $T$ denotes the number of the time step, t is the current time step, and $\alpha_t$ is the noise schedule controlling how fast the data distribution is transformed into a standard Gaussian distribution. As a result, the backward diffusion process in DDPM proves to be 

\begin{align}
x_{t-1} = \frac{1}{\sqrt{\alpha_t}}x_t - \frac{1 - \alpha_t}{\sqrt{\alpha_t (1 - \Pi_{\tau=0}^t \alpha_\tau)}}\mu_\theta^{\epsilon}(x_{t}, t;) + \sigma_t \epsilon
\end{align}

The loss function for DDPM is a denosing autoencoder loss:

\begin{equation}
L_{\theta} = \mathbb{E}_{\boldsymbol{x}_0, \boldsymbol{c}, t} \left\| \boldsymbol{x}_0 - {\boldsymbol{\mu}_{\theta}}^{\boldsymbol{x}_0}(\boldsymbol{x}_t, \boldsymbol{c}, t) \right\|^2
\label{eq:denoise}
\end{equation}

\medskip
\noindent \bfparagraph{Reparameterization.} It is often convenient to reparameterize diffusion models as predicting the \textit{one-step} denoised output 
(called $x_0$-reparameterization) instead of the injected noise (the default $\boldsymbol{\epsilon}$-reparameterization). In DDPM, $x_t = \sqrt{\alpha_t}x_0 + \epsilon_t$ and therefore loss for $\boldsymbol{\epsilon}$-reparameterization is (up to a scale)

\begin{equation}
    L_{\theta} = \mathbb{E}_{\boldsymbol{\epsilon}, \boldsymbol{c}, t} \left\| \boldsymbol{\epsilon} - {\boldsymbol{\mu}_{\theta}}^{\boldsymbol{\epsilon}}(\boldsymbol{x}_t, \boldsymbol{c}, t)\right\|^2
\end{equation}

\medskip
\noindent \bfparagraph{Diffusion Samplers.} When the number of time steps $T$ is large enough, both the forward diffusion process and the backward one can be seen as approximations of their continuous counterparts that can be modeled by stochastic differential equations (SDEs). It is therefore possible to sample from a trained DDPM model with SDE solvers or samplers other than the default DDPM backward diffusion process for better efficiency (at a cost of precision). As an example, DDIM generates samples with

\begin{equation}
x_{t-1} = \sqrt{\alpha_{t-1}} \cdot \left(\frac{x_t - \sqrt{1-\alpha_{t}} \cdot {\boldsymbol{\mu}_{\theta}}^{\boldsymbol{\epsilon}}(x_t, \boldsymbol{c}, t)}{\sqrt{\alpha_t}}\right) + \textit{direction}(x_t) + \tau \boldsymbol{\epsilon}
\label{eq:ddim_sampling}
\end{equation}
where \(\tau\) is a scalar to control the amount of injected noise during the process. Notably, if \(\tau\) is set to 0, DDIM becomes a deterministic sampler (\ie, independent of any noise).

\medskip
\qheading{Text-to-image (T2I) diffusion models.} Different from unconditional diffusion models, T2I diffusion models aim to generate images with optional text prompts. A classical example is Stable Diffusion (SD)~\cite{rombach2021highresolution}, a diffusion model $\mu_\theta(z_t, t, c)$ built with a U-Net architecture and trained on the latent space of a pretrained VAE, in which $c$ is the additional text prompt embedding (typically obtained by CLIP~\cite{radford2021learning}).

\subsection{Diffusion-based Normal Estimator}
\label{sec:overview}

Apart from common multi-modal generation tasks (\eg, text-to-image~\cite{rombach2021highresolution}, text-to-3D~\cite{poole2022dreamfusion}), the pre-trained diffusion models have also proven to have surprisingly good zero-shot performance in several discriminative tasks, such as classification~\cite{li2023your}, and segmentation~\cite{li2023open,tian2024diffuse}. And since image-to-image translation could be considered as a single-modal generation task, different 2D modalities (\eg, image, normal, depth, canny edge) could also be interconverted~\cite{zhang2023adding,ke2023marigold,wang2023context} with the adapted or fine-tuned SD model.

\medskip
\qheading{Normal Estimation with SD}
Since normal estimation can be seen as \textit{translating} an RGB image into a normal map image, the diffusion prior from SD can also be effectively utilized. A straightforward approach is to take the RGB image as the conditioning signal to generate the corresponding normal maps, as in \geowizard~\cite{fu2024geowizard} and \marigold~\cite{ke2023marigold}. 
More specifically, the condition signal is computed by first encoding the RGB input image $\boldsymbol{I}$ into a latent code with a pre-trained VAE encoder, namely $En$, and then, similar to ControlNet~\cite{zhang2023adding}, we transform this latent code $En(\boldsymbol{I})$ through an additional encoder $f_\phi$, into the control signal for the decoder blocks of the U-Net in SD. The decoder blocks of U-Net, which is parameterized by $\theta$, and encoder $f_\phi$ are trained with the following loss (in \(\boldsymbol{\epsilon}\)-reparameterization):

\begin{equation}
L_{\theta, \phi} = \mathbb{E}_{\boldsymbol{\epsilon}, \boldsymbol{c}, \boldsymbol{I}, t} \left\| \boldsymbol{\epsilon} - {\boldsymbol{\mu}_{\theta}}^{\boldsymbol{\epsilon}}(\boldsymbol{x}_t, \boldsymbol{c}, t, f_{\phi}(En(\boldsymbol{I}))) \right\|^2
\label{eq:denoise_control}
\end{equation}
where $\boldsymbol{I}$ is the input image, \(x_t = q(En(N_{\text{gt}}))\) is the latent feature encoded from the ground truth normal map \(N_{\text{gt}}\) at time step \(t\).

During inference, it is straightforward to estimate the normal map for a given RGB image by running anyway sampling algorithm for the trained (conditional) diffusion model. The estimated normal map, though looking sharp, is stochastically generated. We observer that the high variance in the estimated normal maps are typically misaligned with the corresponding input images. While ensemble-like methods can be used (as proposed in \marigold~\cite{ke2023marigold}) to reduce the variance through averaging, the results are still less than satisfactory and the entire ensembling process is quite time-consuming~(see \fref{fig:variance_experiments}).

\medskip
\noindent \bfparagraph{The Variance from the Diffusion Model.}
As argued above, the major issue of diffusion-based normal estimation is the high variance in the diffusion inference procedure. The sources of randomness in diffusion sampling algorithms are mostly 1) the initial Gaussian noise and 2) all intermediate injected Gaussian noises. Thus, we suggest mitigating the variance through a dual-phase inference approach. In the initial phase, a reliable "initial estimate" with high certainty is generated. Subsequently, a second phase of refinement is carried out with a restricted number of diffusion sampling steps, ensuring minimal Gaussian noise injection.

\subsection{You-Only-Sample-Once Normal Initialization}
\label{sec:yoso}

\noindent \bfparagraph{One-step Estimation.} The one-step sampling strategy for normal estimation is firstly introduced in \genpercept~\cite{xu2024diffusion}: no Gaussian noise is introduced, the estimation process is deterministic, but at a cost of overly-smoothing outputs. We instead perform one-step sampling with a Gaussian noise input to balance the sharpness and stability. In mathematical terms, we adopt \initT-parameterization instead of $x_0$-parameterization and reformualte the loss function shown in \cref{eq:denoise_control} to the following one shown in \cref{eq:t+_denoise_control}:

\begin{equation}
L_{\theta, \phi} = \mathbb{E}_{\boldsymbol{x}_{t^{+}}, \boldsymbol{c}, \boldsymbol{I}, t^+} \left\| \boldsymbol{x}_{t^+} - {\boldsymbol{\mu}_{\theta}}^{{x}_{t^+}}(\boldsymbol{x}_\infty, \boldsymbol{c}, t^+, f_{\phi}(En(\boldsymbol{I}))) \right\|^2
\label{eq:t+_denoise_control}
\end{equation}

where \(x_{\infty}\) denotes a noisy sample from the Gaussian distribution resulted from running the forward diffusion process (as in \eref{eq:q_sampling}) with \(t\) approaches infinity and T is set to 1000. Note that we are interested in mapping a distribution from a (standard) Gaussian one to one that corresponds to time $t^+\in (0, T)$, instead to that at time $t=0$.
We call such one-step estimation --- You-Only-Sample-Once (YOSO).
Unfortunately, na\"ively estimating \initT from a Gaussian distribution means learning a many-to-one mapping, which is hard. To address this issue, we propose to use a \DecoupleLoss.

\medskip
\noindent \bfparagraph{\DecoupleLoss.}
We further reduce the variance in the predicted normal maps by training the diffusion model with a regularized loss. 
Instead of penalizing the entropy of the predicted distribution which is generally hard, we take a different path by ``shrinking'' the distribution of predicted normal maps
, $\mu_\theta^{x_t^+}(x_\infty, c, t, f_\phi(En(I)))$, 
to the Dirac delta function 
$\delta(x - \mu_\theta^{x_t^+}(0, c, t, f_\phi(En(I))))$:

\begin{equation}
L_{\theta, \phi} = \begin{cases}
\mathbb{E}_{\boldsymbol{x}_{t^+}, \boldsymbol{c}, \boldsymbol{I}, t^+} \left\| \boldsymbol{x}_{t^+} - {\boldsymbol{\mu}_{\theta}}^{x_{t^+}}(\boldsymbol{x}_\infty, \boldsymbol{c}, t, f_{\phi}(En(\boldsymbol{I}))) \right\|^2, \quad \text{if } p \ge \lambda  \\
\mathbb{E}_{\boldsymbol{x}_{t^+}, \boldsymbol{c}, \boldsymbol{I}, t^+} \left\| \boldsymbol{x}_{t^+} - {\boldsymbol{\mu}_{\theta}}^{x_{t^+}}(0, \boldsymbol{c}, t, f_{\phi}(En(\boldsymbol{I}))) \right\|^2, \quad \text{if } p < \lambda
\end{cases}
\label{eq:decoupled_loss}
\end{equation}
where $p \sim U(0, 1)$, and $\lambda = 0.4$.

\subsection{Semantic-guided Normal Refinement}
\label{sec:sg-refine}
We observe that for subsequent sampling steps that refine the initial normal estimate, the designed image-conditioned diffusion model tends to leverage local instead global information in the RGB image input. However, it is intuitive important not to rely solely on local image information: for instance, to determine the normals for pixels that correspond to a wall, global information is typically much more informative.  We therefore propose to include semantic (and global) features from a pre-trained encoder (for which we use DINO features ~\cite{oquab2024dinov2}) as an auxiliary condition signal.

\smallskip
\noindent \bfparagraph{Architecture of \DINOGuiderAbs.} The entire architecture of \DINOGuiderAbs is depicted in \fref{fig:pipeline}(b), where the image condition branch is denoted by \(f_{\chi}\). It employs a network architecture similar to that in \oneshot except for an extra lightweight semantic-injection network \(g_{\psi}\) that injects the semantic features into the encoder layer of the U-Net in \DINOGuiderAbs (denoted by \DUNET).

\smallskip
\noindent \bfparagraph{Semantic-injection Network.} 
For efficiency,
we implement a lightweight network to feed semantic features into the U-Net. 
Specifically, 
the network employs four conv layers (with 3$\times$3 kernels, 1$\times$1 strides, and channel counts of 16, 32, 64, 128) that are akin to the condition encoder in \cite{zhang2023adding} to align the spatial resolution of DINO features with that of noisy latent features. Given that DINO features typically have a lower resolution than diffusion latent features, for resolution alignment we use FeatUp\cite{fu2024featup} and bi-linear interpolation to upsample DINO features. The noisy latent features are added by the aligned DINO features before being fed into the denoising U-Net.
During training, the network weights are initialized using a Gaussian distribution, except the final projection layer, which is initialized as a zero convolution.

\smallskip
\noindent \bfparagraph{Loss function.} Following the I2VGen-XL~\cite{zhang2023i2vgen}, we reparameterize the \DUNET to the $x_0$-reparameterization. The loss function of \DUNET can be defined as:
\begin{equation}
L_{\theta, \chi, \psi}=\mathbb{E}_{\boldsymbol{x}_0, \boldsymbol{c}, \boldsymbol{I}, \boldsymbol{d}, t}\left\|\boldsymbol{x}_0-{\boldsymbol{\mu}_{\zeta}}^{x_0}\left(\boldsymbol{x}_t, \boldsymbol{c}, t, f_{\chi}(En(\boldsymbol{I})), g_\psi(\boldsymbol{d})\right)\right\|^2
\end{equation}
where $\boldsymbol{d}$ is the processed semantic features extracted from DINO and $t^+\in (0, T)$.

\subsection{Heuristic Denoising Sampling} During inference, 
we apply DDIM to obtain our final normal prediction, as \eref{eq:heri_sampling}. Specifically, the initial normal latent \(x_{t^+}\), predicted from \oneshot, is fed into the solver with 10-step DDIM. Empirically, we set the initial sampling step \(t^{+}\) as 401, which provides an optimal compromise between stability and sharpness.

\begin{equation}
\begin{split}
x_{t-1} &= \sqrt{\alpha_{t-1}}\cdot (\hat{x_0})  + \textit{direction}(x_t)+ \tau\boldsymbol{\epsilon} \\
\hat{x_0} &={\boldsymbol{\mu}_{\zeta}}^{x_0}\left(\boldsymbol{x}_t, \boldsymbol{c}, t, f_{\chi}(En(\boldsymbol{I})), g_\psi(\boldsymbol{d})\right) \\
x_{t^{+}} &= {\boldsymbol{\mu}_{\theta}}^{x_{t^+}}\left(\boldsymbol{x}_\infty, \boldsymbol{c}, t^+, f_{\phi}(En(\boldsymbol{I}))\right)
\label{eq:heri_sampling}
\end{split}
\end{equation}

\section{Experiments}
\label{sec:experiments}
In this section, we compare \modelname with other SOTAs (\ie, DSINE, \marigold, GenPercept and \geowizard) in various real-world datasets.
In addition, an ablation study is conducted to demonstrate the effectiveness of different components, \ie, \oneshot and \DINOGuiderAbs.

\setlength\tabcolsep{0.5em}
\begin{table*}[tb]
\centering
\caption{\textbf{Quantitative comparison on the DTU Dataset~\cite{Jensen2014LargeSM}}. We show the Chamfer distance (Lower is Better). Our method achieves the highest reconstruction accuracy among other normal estimation methods. Different cellcolors refers to \textcolor{orange}{best}, and \textcolor{orange!50}{2nd-best}.}
\vspace{-0.2cm}
\resizebox{.98\textwidth}{!}{
\begin{tabular}{cl|ccccccccccccccc|clc}
\multicolumn{2}{c|}{} & 24 \tdown & 37 \tdown  & 40 \tdown & 55 \tdown  & 63 \tdown  & 65 \tdown & 69 \tdown & 83 \tdown & 97 \tdown & 105 \tdown & 106 \tdown & 110 \tdown & 114 \tdown & 118 \tdown & 122 \tdown & \textbf{Mean} \tdown\\
\shline
\multirow{4}{0.3cm}{\rotatebox[origin=c]{90}{\bf Explicit}}
& 2DGS~\cite{Huang2DGS2024} & \best 0.48 & 0.91 & \tbest \best0.39 & 0.39 & \best 1.01 & \sbest 0.83 & 0.81 & 1.36 & \sbest1.27 & \best0.76 & 0.70 & 1.40 & \sbest0.40 & 0.76 & 0.52 & 0.80 \\
& 2DGS + DSINE\cite{bae2024dsine} & 0.62 & 0.76 & 0.49 & \sbest0.38 & 1.20 & 1.04 & 0.68 & 1.34 & 1.35 & \sbest0.76 & \sbest0.61 & 0.83 & 0.42 & \sbest0.57 & \sbest0.44 & 0.76 \\
& 2DGS + \geowizard\cite{fu2024geowizard} & 0.54 & \sbest0.75 & 0.43 & \sbest0.38 & \sbest1.15 & \best0.80 & \sbest0.66 & \best1.28 & 1.47 & 0.80 & \sbest0.61 & \sbest0.81 & \sbest0.40 & 0.59 & 0.50 & \sbest0.75 \\
& 2DGS + Ours & \sbest0.51 & \best0.72 & \sbest0.41 & \best0.38 & 1.18 & 0.86 & \best0.61 & \sbest1.29 & \best1.09 & 0.84 & \best0.59 & \best0.79 & \best0.36 & \best0.54 & \best0.43 & \best0.70 \\
\end{tabular}}
\label{tab:dtu_result}
\end{table*}

\subsection{Experimental Setup}
\zheading{Datasets} Following \geowizard~\cite{fu2024geowizard}, our model is trained on a comprehensive dataset of high-resolution images and ground truth normals rendered from synthetic scenes across three categories: 25,463 samples from HyperSim \cite{roberts2021hypersim} and 50,884 samples from Replica \cite{straub2019replica} for indoor environments; 76,048 samples from 3D Ken Burns \cite{Niklaus_TOG_2019} and 39,630 synthetic city images from MatrixCity \cite{matrixcity}; and 85,997 background-free 3D objects from Objaverse \cite{objaverse}. Most of the data is photorealistically rendered using Blender and Unreal Engine, totaling over 250,000 image-normal pairs.

\qheading{Implementation} We fine-tune the Stable Diffusion V2.1~\footnote{\href{https://huggingface.co/stabilityai/stable-diffusion-2-1}{hf.co/stabilityai/stable-diffusion-2-1}} using the AdamW optimizer\cite{loshchilov2019decoupled} with a fixed learning rate of 3e-5. Please check out more implementation details in \supmat's \cref{sup-sec: implementation}.

\qheading{Metrics} For evaluation, we follow the metrics outlined in DSINE~\cite{bae2024dsine} and calculate the angular error between the estimated and ground truth normal maps. We report both the mean and median angular errors, with lower values indicating better accuracy. Additionally, we measure the percentage of pixels with an angular error below specified thresholds of \(11.25^\circ\), \(22.5^\circ\), and \(30.0^\circ\), where higher percentages reflect superior performance.

\subsection{Comparison to the state-of-the-art}
We choose DSINE \cite{bae2024dsine}, Marigold \cite{ke2023marigold} (normal version\footnote{\href{https://huggingface.co/prs-eth/marigold-normals-lcm-v0-1}{hf.co/prs-eth/marigold-normals-lcm-v0-1}}, denote as Marigold \textsuperscript{\textdagger}), GenPercept~\cite{xu2024diffusion} and GeoWizard~\cite{fu2024geowizard} for comparison. DSINE is the SOTA method among all regression-based methods and \geowizard is the SOTA among all existing diffusion-based ones. 
Due to the unavailability of DSINE's training data, we retrained the model using the provided code and our dataset. Nonetheless, our retrained model underperformed compared to the original released version, so we decided to use the original model for our evaluation.
For \geowizard, since the training code is not available, we utilized the pre-released model~\footnote{\href{https://huggingface.co/lemonaddie/Geowizard}{hf.co/lemonaddie/Geowizard}} for our evaluations. We consider this approach fair because we use the same training dataset.

\begin{table}[tb]
\footnotesize
\caption{
        \textbf{Quantitative evaluation}. Here we compare with DSINE~\cite{bae2024dsine}, \marigold\textsuperscript{\textdagger}~\cite{ke2023marigold}, and \geowizard~\cite{fu2024geowizard}, another two diffusion-based normal estimators, on four indoor benchmarks. Different cellcolors refers to \textcolor{orange}{best}, and \textcolor{orange!50}{2nd-best}.
    } 
\begin{tabular}{l|cc|ccc}
Method & mean \tdown & med\tdown & {\scriptsize $11.25^{\circ}$}\tup & {\scriptsize $22.5^{\circ}$}\tup & {\scriptsize $30^{\circ}$}\tup \\
\shline
\multicolumn{6}{c}{NYUv2~\cite{Silberman2012IndoorSA}} \\
\hline
\geowizard & 20.363 & 11.898 & 46.954 & 73.787 & 80.804  \\
\marigold\textsuperscript{\textdagger} & 20.864 & 11.134 & 50.457 &  73.003 & 79.332\\
GenPercept & 20.896 & 11.516 & 50.712 & 73.037 & 79.216 \\
DSINE & \best 18.610 & \best 9.885  & \best 56.132 & \best 76.944 & \best 82.606 \\
Ours & \sbest 19.707 & \sbest 10.527 & \sbest 53.042 & \sbest 75.889 & \sbest 81.723 \\
\hline
\multicolumn{6}{c}{ScanNet~\cite{dai2017scannet}}\\
\hline
\geowizard & 21.439 &  13.930 & 37.080 & 71.653 & 79.712  \\
\marigold\textsuperscript{\textdagger} & 21.284 & 12.268 & 45.649  & 72.666 & 79.045 \\
GenPercept & 20.652 & 10.502 & 53.017 & 74.470 & 80.364 \\
DSINE & \sbest 18.610 & \best 9.885  & \best 56.132 & \sbest 76.944 & \sbest 82.606 \\
Ours & \best 18.098 & \sbest 10.097 & \sbest 56.007 & \best 78.776 & \best 84.115  \\
\hline
\multicolumn{6}{c}{iBims-1~\cite{koch2018evaluation}}\\
\hline
\geowizard & 19.748 & 9.702 &  58.427 & 77.616 & 81.575 \\
\marigold\textsuperscript{\textdagger}& \sbest 18.463 & 8.442  & \sbest 64.727 & \sbest 79.559 & \sbest 83.199\\
GenPercept & 18.600 & 8.293 &  64.697 & 79.329 & 82.978 \\
DSINE & 18.773 & \sbest 8.258 & 64.131 & 78.570 & 82.160  \\
Ours & \best 17.248 & \best 8.057 & \best 66.655 & \best 81.134 & \best 84.632 \\
\hline
\multicolumn{6}{c}{DIODE-indoor~\cite{vasiljevic2019diode}} \\
\hline
\geowizard & 19.371 & 15.408 & 30.551 & 75.426 & 86.357 \\
\marigold\textsuperscript{\textdagger}  & \sbest 16.671 & \sbest 12.084 & \sbest 45.776 & \sbest 82.076 & \sbest 89.879\\
GenPercept & 18.348 & 13.367 & 39.178 & 79.819 & 88.551 \\
DSINE & 18.453 & 13.871 & 36.274 & 77.527 & 86.976 \\
Ours & \best 13.701 & \best 9.460  & \best 63.447 & \best 86.309 & \best 92.107 \\
\end{tabular}
\label{table:benchmark}
\end{table}

\begin{figure*}[htb] \centering
    \includegraphics[width=0.8\linewidth]{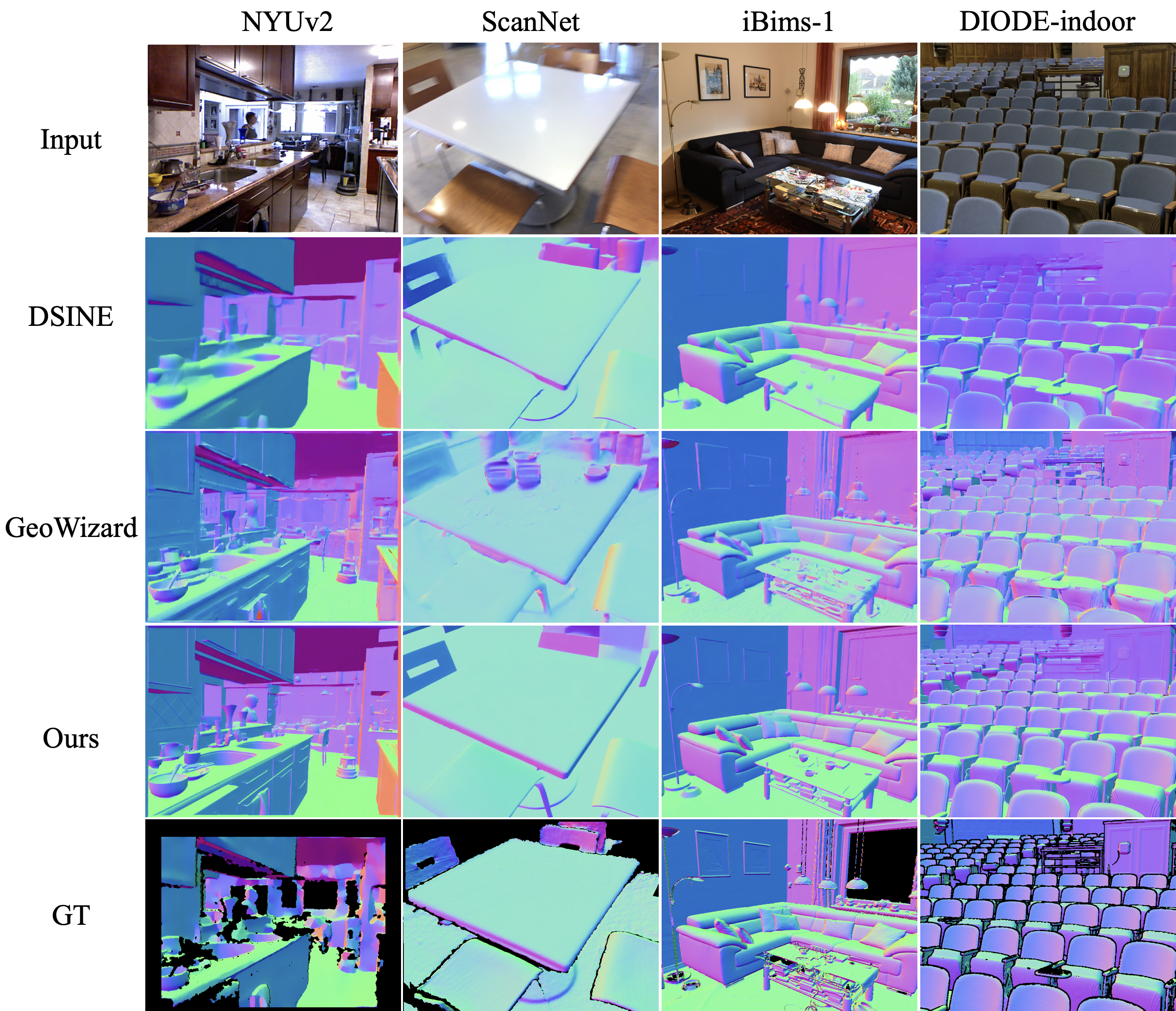}
    \caption{Qualitative comparison of different methods on NYUv2\cite{Silberman2012IndoorSA}, ScanNet\cite{dai2017scannet}, iBims-1\cite{koch2018evaluation}, DIODE-indoor\cite{vasiljevic2019diode} datasets. \modelname outperforms other related works in terms of accuracy and sharpness.
    } 
    \label{fig:benchmark}
\end{figure*}

The testing data for evaluation includes the challenging DIODE-indoor \cite{vasiljevic2019diode}, iBims \cite{koch2018evaluation}, ScanNetV2 \cite{dai2017scannet}, and NYUv2 \cite{Silberman2012IndoorSA} datasets. As presented in \tref{table:benchmark}, our method achieves superior performance across iBims, ScanNetV2, and DIODE-indoor by a large margin. On NYUv2, our method is slightly inferior to DSINE. We argue that both Scannet and NYUV2 are captured using low-quality sensors, thus their GT normal are not accurate, which is also mentioned in~\geowizard~\cite{fu2024geowizard}). \Cref{fig:benchmark} shows the qualitative comparisons on challenging scenarios, which demonstrates the accuracy and sharpness of \modelname.

\begin{figure}[htb] \centering
        \begin{tabularx}{\linewidth}{
 >{\centering\arraybackslash}X
 >{\centering\arraybackslash}X
}
(a) Output Variance Analysis& (b) Inference Time Analysis
 \end{tabularx}
    \includegraphics[trim=000mm 012mm 000mm 000mm, clip=true, width=8cm]{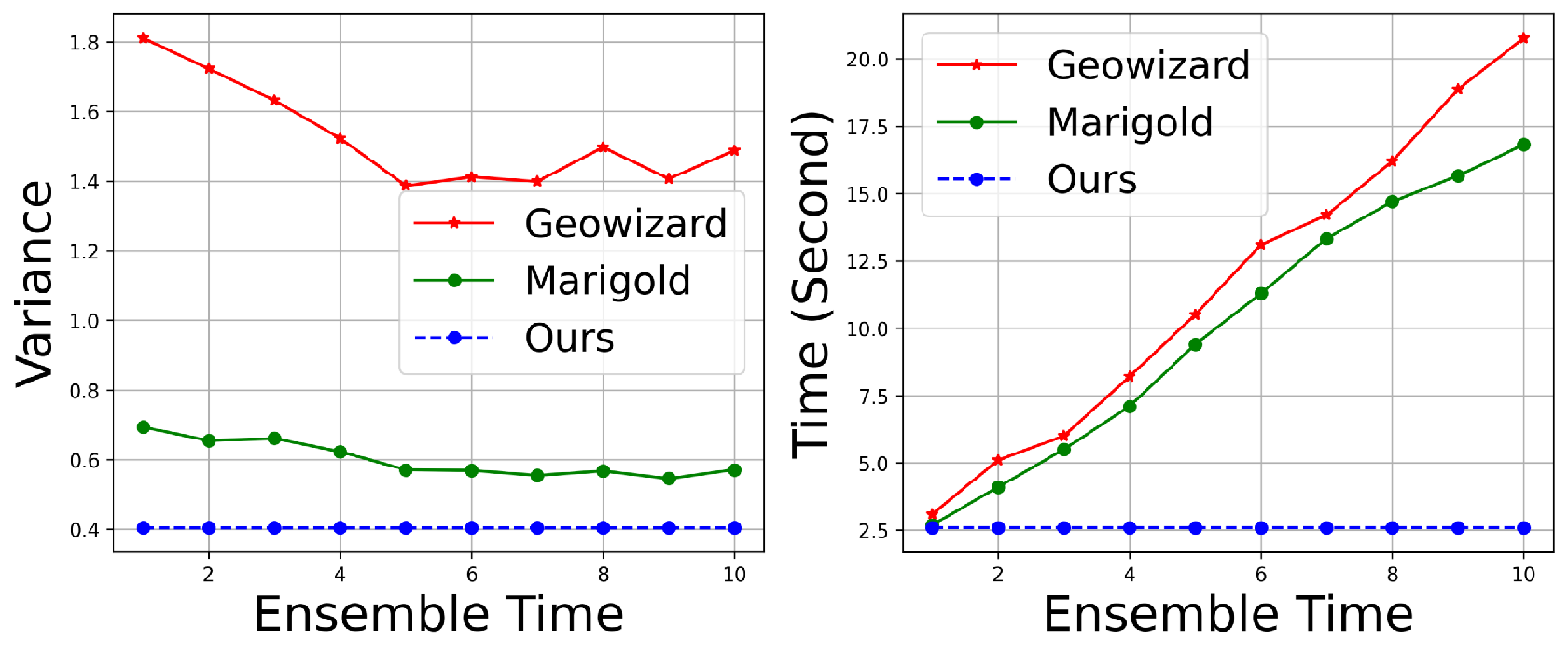}
\begin{tabularx}{\linewidth}{
 >{\centering\arraybackslash}X
 >{\centering\arraybackslash}X
}
Ensemble Times& Ensemble Times
 \end{tabularx}
    \caption{The comparison of output variance and inference time between our method, GeoWizard, and \marigold. The left plot shows the output variance over ensemble time, while the right plot displays the inference time (including ensembling). It is important to note that our method does not employ the ensemble strategy and only requires a single forward pass.
    } 
    \label{fig:variance_experiments}
\end{figure}

\Cref{fig:variance_experiments} compares the inference variance and time between our method and \geowizard on the DIODE-indoor dataset. Specifically, we estimate each image 10 times using different initialization seeds, allowing us to calculate the variance for each individual image. We then calculated the overall variance for each model by averaging these values across the entire dataset. 
As shown in \cref{fig:variance_experiments}~(a), \geowizard employs an ensemble strategy to reduce the variance of the output. However, our approach significantly decreases the output variance (0.410 vs. 1.370) without introducing any ensemble strategies. Furthermore, the ensemble strategy compromises speed to achieve a lower variance. \Cref{fig:variance_experiments}~(b) shows that \geowizard samples five times (approximately 10 seconds) to reach a variance of 1.370, while our method achieves a variance of 0.410 within 3 secs. The inference speed was tested on a single A100 GPU.

\subsection{Ablation study}

\begin{table}[t!]
\footnotesize
\caption{\textbf{Ablation Studies.} Different cellcolors refers to \textcolor{orange}{best} and \textcolor{orange!50}{2nd-best}.}
\begin{tabular}{l|cc|ccc}
& mean \tdown & med\tdown & {\scriptsize $11.25^{\circ}$}\tup & {\scriptsize $22.5^{\circ}$}\tup & {\scriptsize $30^{\circ}$}\tup \\
\shline
\multicolumn{6}{c}{NYUv2~\cite{Silberman2012IndoorSA}}\\
\hline
Ours & 19.707 & 10.527 & 53.042 & 75.889 & 81.723 \\
\oneshot Only & \sbest 18.917 & \sbest 10.509 & \sbest 53.074 & \sbest 76.008 & \sbest 82.524 \\
Ours w/o DINO &  19.739 &  10.536 &  52.999 &  75.833 &  81.667\\
\hline
DSINE & \best 18.610 & \best 9.885  & \best 56.132 & \best 76.944 & \best 82.606 \\
\DINOGuiderAbs + DSINE  & 19.869 & 10.548 & 52.952 & 75.738 & 81.575\\

\hline
\multicolumn{6}{c}{ScanNet~\cite{dai2017scannet}}\\
\hline
Ours & \sbest 18.098 & 10.097 & 56.007 & \sbest 78.776 & \sbest 84.115  \\
\oneshot Only & \best 17.679  & \best 9.860  & \best 57.220 & \best 78.823 & \best 84.331 \\
Ours w/o DINO & 19.326 & 11.626 & 48.115 &  77.438 &  83.575\\
\hline
DSINE & 18.610 & \sbest 9.885  & \sbest 56.132 & 76.944 & 82.606 \\
\DINOGuiderAbs + DSINE &  19.118 &  10.221 &  54.789 & 77.115 & 82.568\\

\hline
\multicolumn{6}{c}{iBims-1~\cite{koch2018evaluation}}\\
\hline
Ours & \best 17.248 & \best 8.057  & \best 66.655 & \best 81.134 & \best 84.632 \\
\oneshot Only & \sbest 17.695 &  8.431 &   63.635 & 80.212 &  84.034  \\
Ours w/o DINO & 18.234 & 8.875 & 62.172 &  80.417 & \sbest 84.347\\
\hline
DSINE & 18.773 & 8.258 & 64.131 & 78.570 & 82.160  \\
\DINOGuiderAbs + DSINE &  17.877 & \sbest 8.069 &  \sbest 66.589 & \sbest 80.630 & 83.957\\

\hline
\multicolumn{6}{c}{DIODE-indoor~\cite{vasiljevic2019diode}} \\
\hline
Ours & \best 13.701 & \best 9.460  & \best 63.447  & \best 88.223 & \best 92.107 \\
\oneshot Only &  17.122 &  13.787 & 32.950 &  83.385 &  89.884 \\
Ours w/o DINO &  15.611 &  11.912 &  45.801 & \sbest 86.563 & \sbest 91.843 \\
\hline
DSINE & 18.453 & 13.871 & 36.274 & 77.527 & 86.976 \\
\DINOGuiderAbs + DSINE & \sbest  14.752 & \sbest  10.139 &  \sbest 58.221 & 86.455 & 90.888 \\

\end{tabular}
\label{table:ablation}
\end{table}

\begin{table}[tb]
\footnotesize
\caption{Ablation study of the effectiveness of \DecoupleLoss. Best results are \textcolor{orange}{highlighted}.}
\begin{tabular}{l|cc|ccc}
Ablation & Mean \tdown & Med\tdown & { $11.25^{\circ}$}\tup & { $22.5^{\circ}$}\tup & { $30^{\circ}$}\tup \\
\shline
\multicolumn{6}{c}{DIODE-indoor~\cite{vasiljevic2019diode}}\\
\hline
w/o \DecoupleLoss &  18.624 & 14.237 & \best 37.504 & 76.569 & 87.740 \\
w/ \DecoupleLoss & \best 17.122 & \best 13.787 & 32.950 &  \best 83.385 &  \best 89.884 \\
\hline
\multicolumn{6}{c}{iBims-1~\cite{koch2018evaluation}} \\
\hline
w/o \DecoupleLoss & 18.552 & 9.049 & 61.791 & 79.077  &81.852 \\
w/ \DecoupleLoss &  \best 17.695 & \best 8.431 & \best 63.635 & \best 80.212 & \best 84.034 \\
\end{tabular}
\label{tab:decoupled_loss}
\end{table}

\begin{figure*}[tb] \centering
    \includegraphics[width=\textwidth]{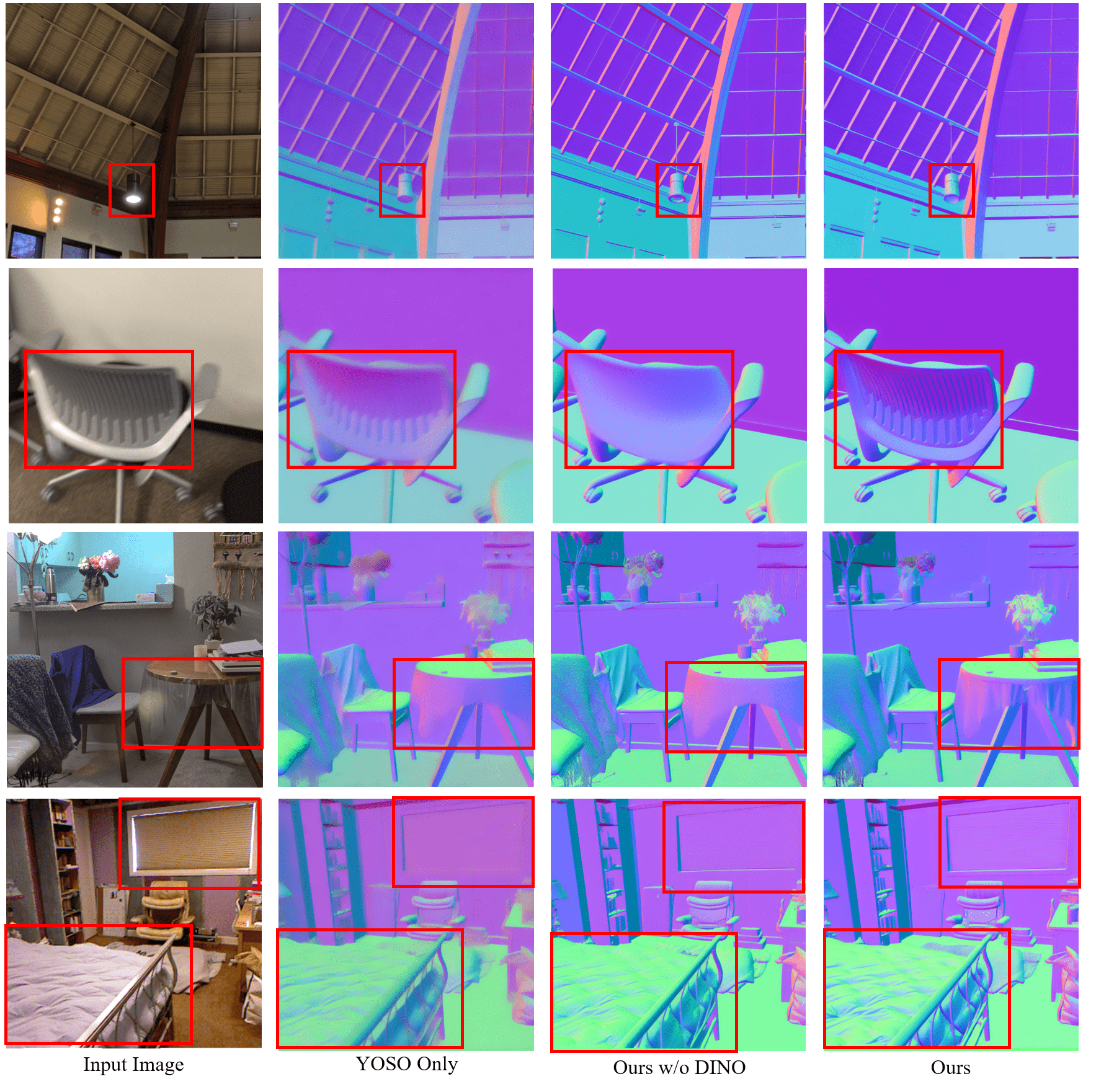}
    \caption{
        \textbf{Qualitative Ablation Study.}
        \emph{\oneshot} can produce relatively sharp surface normal estimations with only a single-step sampling; however, its results still lack sufficient details. After refinement by \DINOGuiderAbs, the predicted surface normals become significantly sharper, as illustrated by the comparison between the third and fourth columns in the figure. This comparison highlights the impact of semantic features on \DINOGuiderAbs's performance. Specifically, the first row demonstrates how using DINO features assists the network in mitigating the effects of lighting on normal estimation. The second row indicates that DINO features enable effective structural modeling, enhancing the consistency of the normal output. Furthermore, the third row shows that DINO features improve the network's ability to understand materials, \eg, plastic material.
    } 
    \label{fig:ablations_DINO}
    \vfill\null
\end{figure*}

We conduct ablation studies to analyze the contribution of each component in our framework across four datasets: NYUv2, ScanNet, iBims-1, and DIODE-indoor. Both quantitative and qualitative results are summarized in \cref{table:ablation} and \cref{fig:ablations_DINO}.

\qheading{Ablation on SG-DRN} We first evaluated the refinement step -- SG-DRN. We refer to the method without the refinement pipeline as \emph{\oneshot Only}. As shown in \cref{table:ablation}, there is a performance degradation on both the iBims-1 and DIODE-indoor datasets, highlighting the critical role of the SG-DRN refinement module in improving normal estimation accuracy. Notably, since NYUv2 and ScanNet feature smooth GT normals, and the prediction normals by \emph{\oneshot Only} are relatively smooth as well, the quantitative performance of \emph{\oneshot Only} even surpasses that of the full version with the refinement process. However, this is not the case when examining the qualitative results (see \supmat's \cref{supp: qualitative-yoso-only-with-ours}). Furthermore, we also evaluate the DSINE with SG-DRN module, refered as \emph{SG-DRN+DSINE}, the results on DIODE-indoor and iBIMS-1 datasets also justify the effectiveness of multi-step refinement.

\medskip
\qheading{YOSO Normal Initialization} Next, we investigate the effect of the YOSO initialization . To do this, we tried an alternative to use the output of the DSINE method instead of our YOSO as the initialization, which is termed as \emph{SG-DRN $+$ DSINE}. The results on the DIODE-indoor dataset reveal that using DSINE's initialization leads to an increase in mean angle error from 13.701° to 18.453°. This verifies that the necessity of our YOSO initialization.

\medskip
\qheading{Ablation on Semantic feature extractor} There are alternatives for extracting semantic features. We denote the one replacing DINO extractor with a standard ResNet-50 backbone as \emph{Ours w/o DINO}, with which, the performance decreases across all datasets, validating that the superiority of DINO visual representation to be the semantic guidance for normal estimation. The most significant drop is observed on the DIODE-Indoor dataset, where the mean angle error rises from 13.701° to 15.611°. Qualitative comparisons in \fref{fig:ablations_DINO} further verifies the usefulness of DINO features.

\medskip
\qheading{Effects of \DecoupleLoss} \Tref{tab:decoupled_loss} illustrates that our proposed \DecoupleLoss can effectively mitigate the difficulty of learning many-to-one mapping, improving overall metrics on both DIODE-indoor and iBims-1 benchmark.

\section{Applications}
\subsection{Multi-view Surface Reconstruction}
\begin{figure}[t] \centering
    \includegraphics[width=\linewidth]{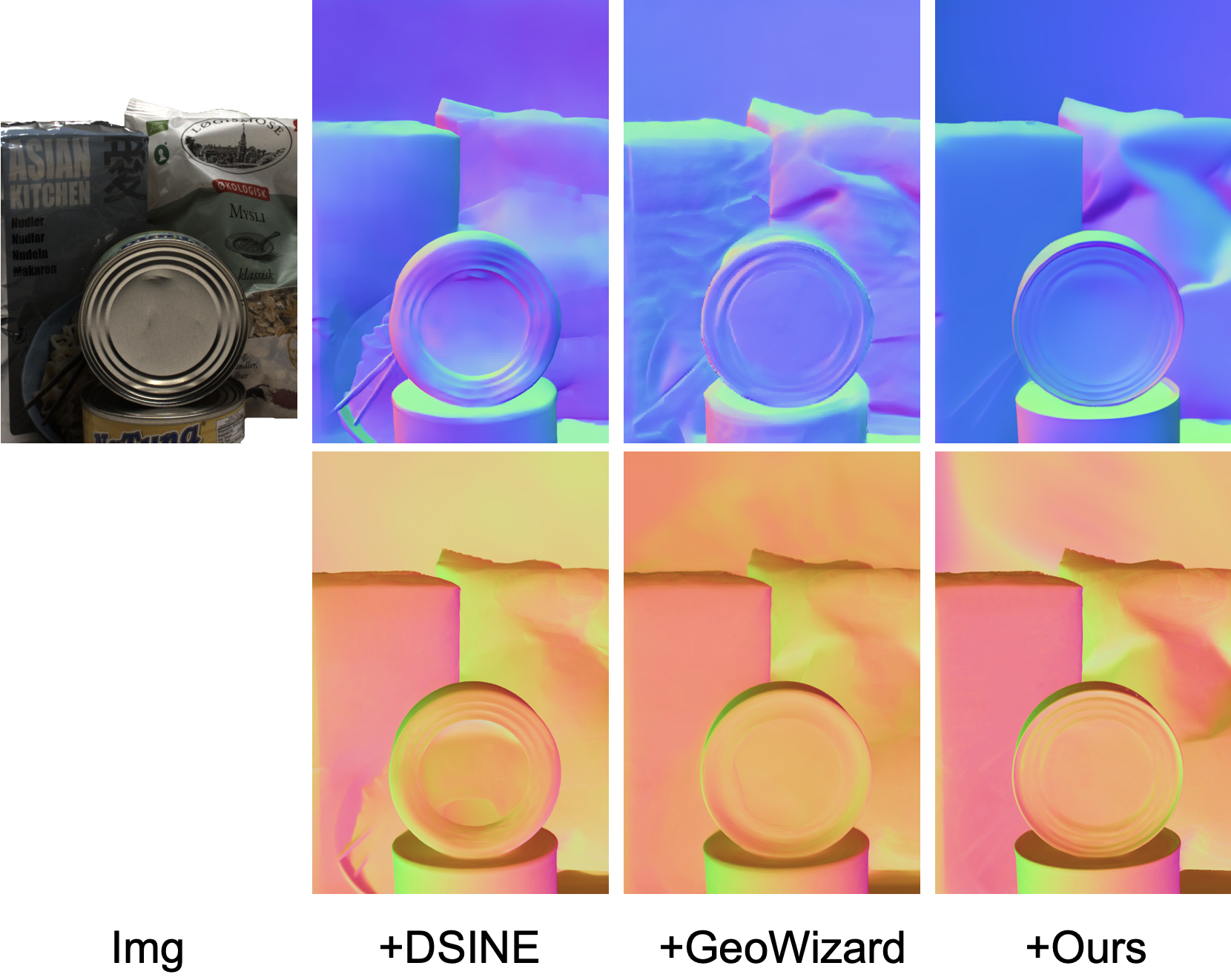}
    \caption{Qualitative comparison on DTU\cite{Jensen2014LargeSM}dataset. The first row displays the input images and estimated normal maps. The second row visualizes the rendered world-space normal maps after the reconstruction.
    } 
    \label{fig:application_object_reconstruction}
\end{figure}

Accurate normal estimation is crucial for faithful surface reconstructions, especially for non-Lambertian surfaces (\cref{fig:application_object_reconstruction}). We leverage our generated normal maps to regularize the surface reconstruction pipeline following 2DGS~\cite{Huang2DGS2024}. Quantitative results on DTU (\cref{tab:dtu_result}) show our method achieves the lowest mean Chamfer distance among compared techniques, highlighting the significant impact of our accurate normal estimates.

\subsection{Monocular Surface Reconsturction}
\begin{figure}[t] \centering
    \includegraphics[width=\linewidth]{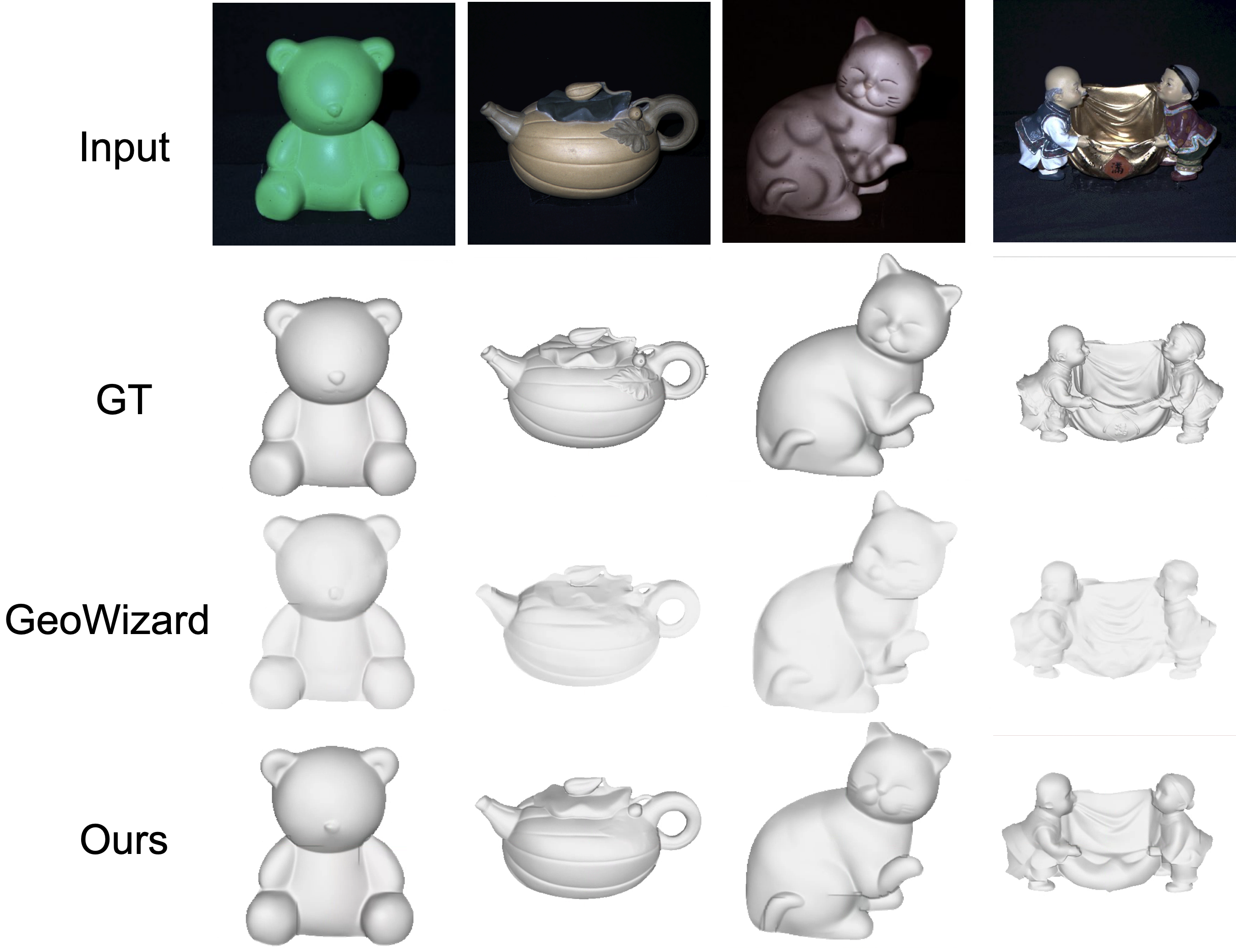}
    \caption{Qualitative comparison on DiLiGenT\cite{Shi2019ABD} dataset.
    } 
    \label{fig:application_normal_intergration}
\end{figure}

Our high-fidelity normal estimation also benefits monocular surface reconstruction via normal field integration, like Bilateral Normal Integration (BiNI)~\cite{cao2022bilateral}. We compare monocular geometric regularization from different methods on 80 DiLiGenT samples with ground-truth normals. Table \ref{tab:bini_result} reports our method significantly improves Normal RMSE, Mean Angle Error (by ~20\%), and Depth Mean Angle Error over previous methods, demonstrating robust normal estimation across lighting conditions. Fig. \ref{fig:application_normal_intergration} visualizes extracted mesh comparisons against GT and GeoWizard, showing our method faithfully recovers intricate geometric structures.

\begin{table}[H]
\footnotesize
\caption{Quantitative evaluation on the DiLiGenT~\cite{Shi2019ABD} dataset for monocular surface reconsturction application. Different cellcolors refers to \textcolor{orange}{best}, and \textcolor{orange!50}{2nd-best}.}
\begin{tabular}{l|ccc}
Method & N-RMSE $\downarrow$ & MAE $\downarrow$ & D-RMSE $\downarrow$\\
\shline
DSINE\cite{bae2024dsine} & 0.50 & \sbest22.53  & 0.0053 \\
GeoWizard\cite{fu2024geowizard} & \sbest0.49 & 24.51 & \sbest0.0048 \\
Ours & \best0.41 & \best18.78  & \best0.0044\\
\end{tabular}
\label{tab:bini_result}
\end{table}

\subsection{Normal Enhancement}
\begin{figure*}[htb] \centering
    \includegraphics[width=\linewidth]{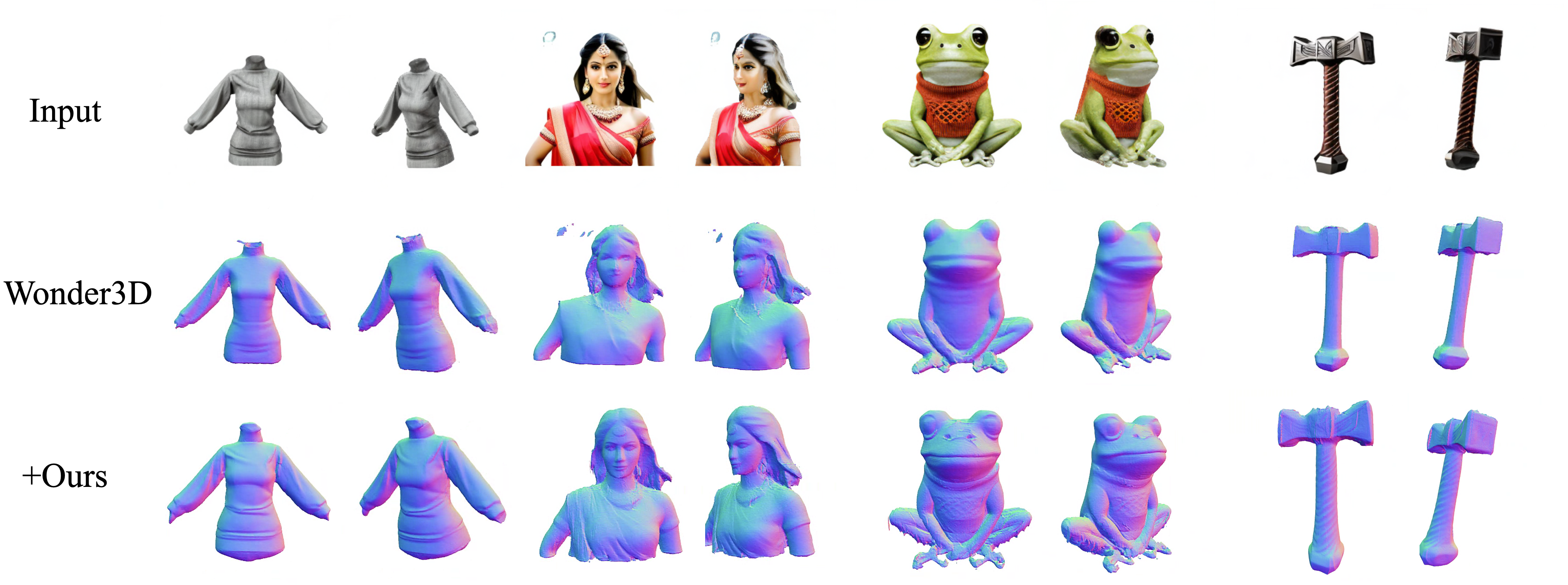}
    \caption{Comparison of geometric surface normals for different scenes. The surface normals are rendered from the reconstructed 3D mesh models.} 
    \label{fig:application_normal_enhancement}
    \vspace{-5mm}
\end{figure*}

Recent generative AI advances enable 3D content creation by finetuning pre-trained 2D diffusion models to predict multi-view normal maps~\cite{lu2024direct25,zheng2024mvd2,qiu2023richdreamer,long2023wonder3d}, which are then fused into 3D models. However, existing methods produce low-resolution and over-smooth outputs lacking fine details. To improve it, we apply our method to Wonder3D~\cite{long2023wonder3d} to improve the detail of the generated multi-view normal maps and the resulting 3D shapes. We upsample the multi-view images using bilinear upsampling and the low-res normal maps to initialize $x_t$, leveraging their multi-view consistency. Our SG-DRN then refines the upsampled normal maps to recover finer details. Following Wonder3D~\cite{long2023wonder3d}, we train a NeuS~\cite{wang2021neus} per object using the refined normal maps and extract high-res meshes. \Fref{fig:application_normal_enhancement} shows our method significantly improves the detail of the generated 3D objects compared to the original one.

\section{Conclusion}

We present \modelname, which tailors the diffusion priors for monocular normal estimation. Unlike prior diffusion-based works, we prioritize enhancing estimation stability by reducing inherent diffusion stochasticity. Our approach, a coarse-to-fine strategy, hinges on the belief that a reliable initial guess combined with a semantic-guided refinement process is crucial for balancing the ``stability vs. sharpness'' trade-off. This is validated by multiple indoor benchmarks, and various real-world applications (check our \vid for more details). Some failure cases are in~\supmat's \cref{sup-sec:failure}. While our focus is on normal estimation, we believe our methodology and the identified trade-off will also benefit other related fields, including but not limited to depth estimation and various perception tasks (\eg, detection, segmentation, \etc). To democratize this, we will make our code and models publicly available, only for research purpose.

\bigskip
\noindent\rule[3.0ex]{\linewidth}{1.5pt}

\qheading{Acknowledgments} 
We thank \textit{Guanying Chen} and \textit{Zhen Liu} for proofreading, \textit{Zhen Liu} and \textit{Xu Cao} for fruitful discussions.

\clearpage
\bibliographystyle{ACM-Reference-Format}
\bibliography{reference}

\clearpage
\appendix

\renewcommand{\thefigure}{R.\arabic{figure}}
\renewcommand{\thetable}{R.\arabic{table}}
\renewcommand{\theequation}{S.\arabic{equation}}

\setcounter{figure}{0}
\setcounter{table}{0}
\setcounter{equation}{0}

\begin{subappendices}
\renewcommand{\thesection}{\Alph{section}}%
\label{sec:appendix}

\section{More details about implementation} 
\label{sup-sec: implementation}

We fine-tune the pre-trained Stable Diffusion V2.1 \cite{Rombach_2022_CVPR} using the AdamW optimizer\cite{loshchilov2019decoupled} with a fixed learning rate of 3e-5. To enhance the robustness of our method against exposure, we incorporate exposure augmentation. Furthermore, we transform all input maps to the range [-1, 1] to align with the VAE's expected input range. During training, we employ random crops with varying aspect ratios and pad the images to a fixed box resolution using black padding. Our training process involves two stages: first, we pre-train our network with a resolution of 512x512 using a batch size of 64 for around 20,000 steps. Subsequently, we fine-tune the model on a 768x768 resolution with a batch size of 32 for 10,000 steps. The entire training process takes approximately one day on four A100 GPUs. Notably, both \oneshot and \DINOGuiderAbs employ the same training strategy.

\begin{figure}[htb] 
    \includegraphics[width=0.9\linewidth]{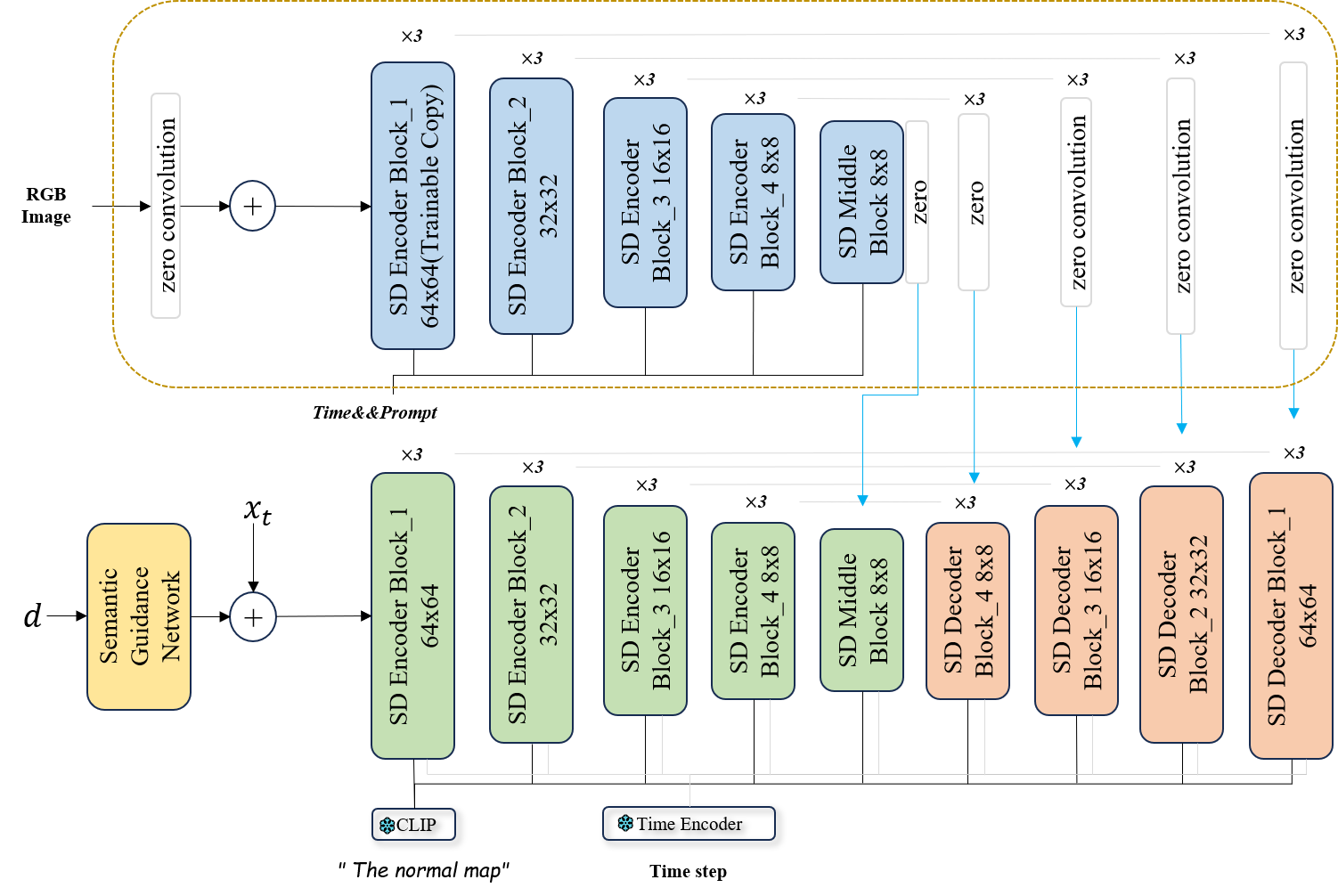}
    \caption{The details of the architecture of our U-Net. } 
    \label{supp: archit}
\end{figure}

\section{The architecture of U-Net in both stages}
\label{sup-sec: arch}
Our structure maintains most building blocks of ControlNet \cite{zhang2023adding} with several modifications for normal estimation(we show the second stage here). As depicted in Figure. \ref{supp: archit}, we use a fixed text prompt ``The Normal Map" in both the training and testing phases and add a semantic-guider network to encode DINO features. The encoded feature is further added with the output of the YOSO stage to act as input to the SG-DRN. The semantic guider is a simple stacking of 2D convolutions for obtain features, following by Featup~\cite{fu2024featup} and bi-linear interpolation to upsample their resolution to the same shape as the YOSO output.

\begin{figure}[htb] 
    \includegraphics[width=\linewidth]{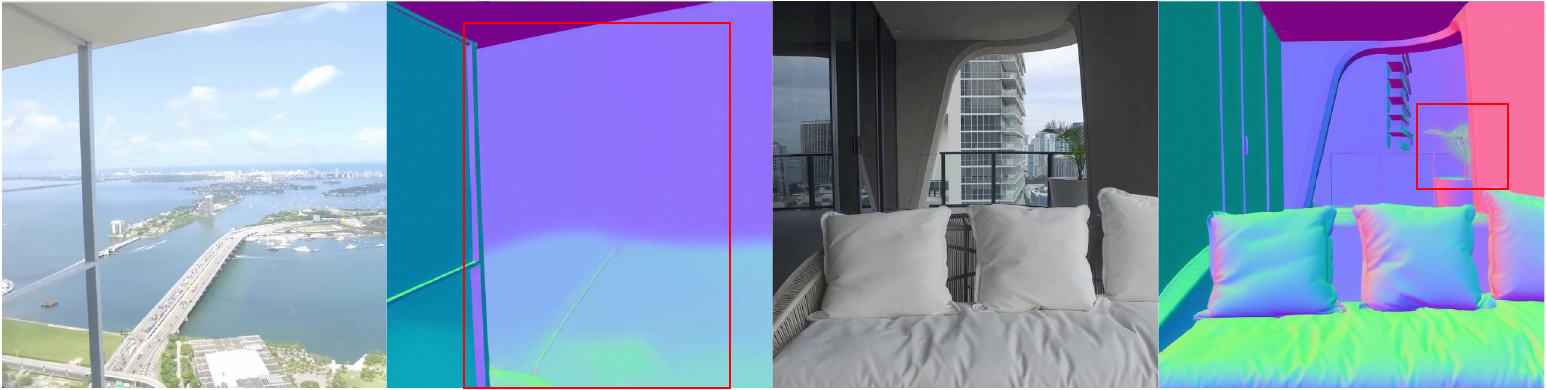}
    \caption{Typical bad cases generated by StabelNormal.} 
    \label{supp: badcases}
\end{figure}

\section{Failure Case} 
\label{sup-sec:failure}

While StableNormal can produce sharp and stable normal estimation under most circumstances, it may also fail in some extreme cases like all data-driven methods. As depicted in Figure. \ref{supp: badcases}, StableNormal could partially output the normal of things behind the transparent objects(\textbf{Left}) and output a similar color(green) for plants in images(\textbf{Right}) regardless of the complex normal directions on the surface of plants. This is due to the inductive bias introduced by our training dataset(Lack of data including outdoor scenes and plants), which could be solved in the future by adding more simulating renderings.

\section{More qualitative analysis of YOSO}

\begin{figure}[htb] 
    \includegraphics[width=\linewidth]{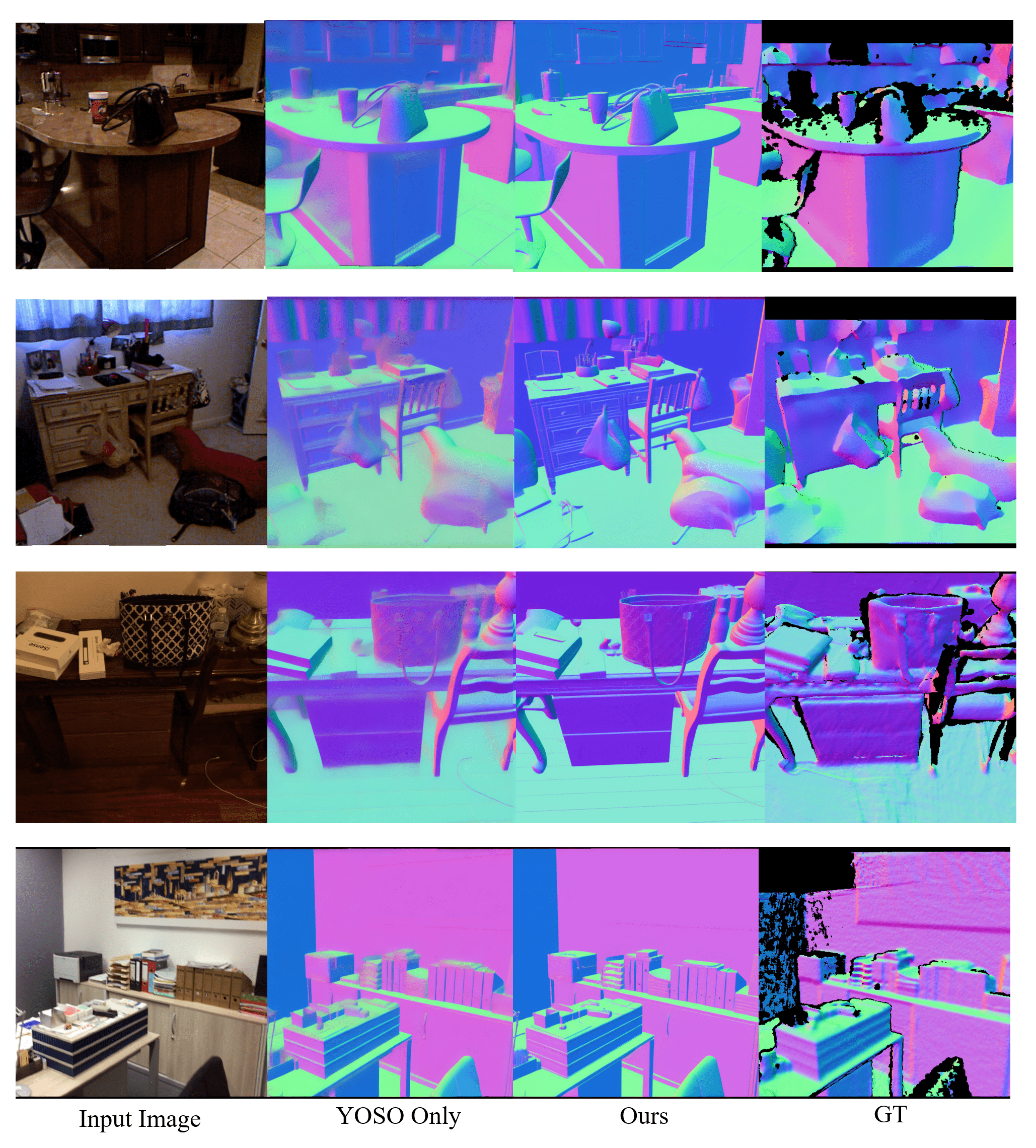}
    \caption{The qualitative comparison results between YOSO Only and Ours on both  NYUv2 and ScanNet dataset.} 
    \label{supp: qualitative-yoso-only-with-ours}
\end{figure}

Although our method predicts sharper and more accurate normals compared to YOSO Only, the qualitative results appear worse than those of YOSO Only because the ground truth normal maps of both NYUv2 and ScanNet are smoother and less detailed (see \fref{supp: qualitative-yoso-only-with-ours}).

\section{More qualitative comparisons}
\begin{figure*}[htb] 
    \includegraphics[width=0.9\linewidth]{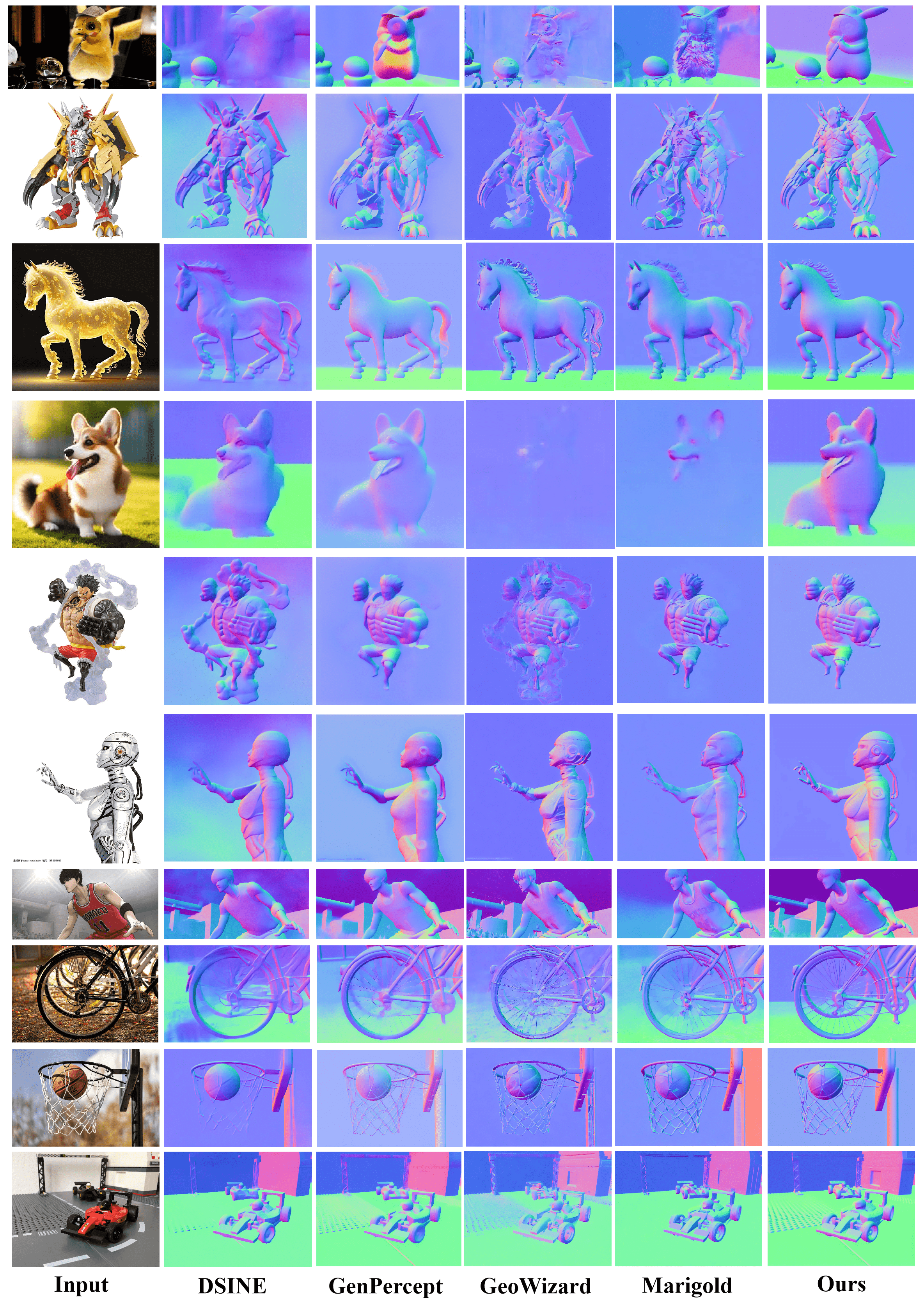}
    \caption{More qualitative results~(Part I). From the left to the right are results from DSINE~\cite{bae2024dsine}, GenPercept~\cite{xu2024diffusion}, GeoWizard~\cite{fu2024geowizard}, Marigold~\cite{ke2023repurposing} and \modelname respectively.} 
    \label{supp:more-qualitative1}
\end{figure*}

\begin{figure*}[htb] 
    \includegraphics[width=0.8\linewidth]{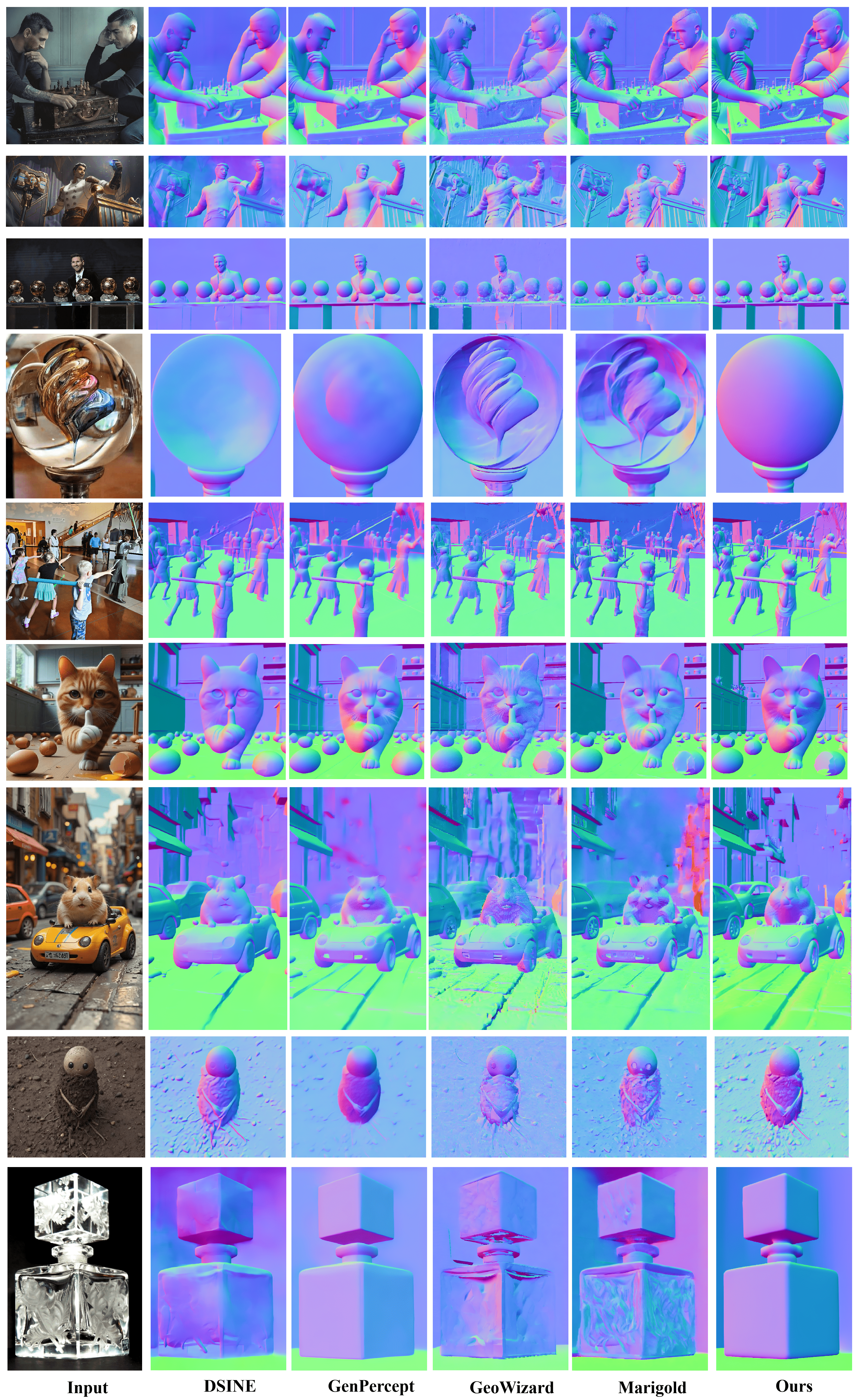}
    \caption{More qualitative results~(Part II). From the left to the right are results from DSINE~\cite{bae2024dsine}, GenPercept~\cite{xu2024diffusion}, GeoWizard~\cite{fu2024geowizard}, Marigold~\cite{ke2023repurposing} and \modelname respectively.} 
    \label{supp:more-qualitative2}
\end{figure*}

\begin{figure*}[htb] 
    \includegraphics[width=0.9\linewidth]{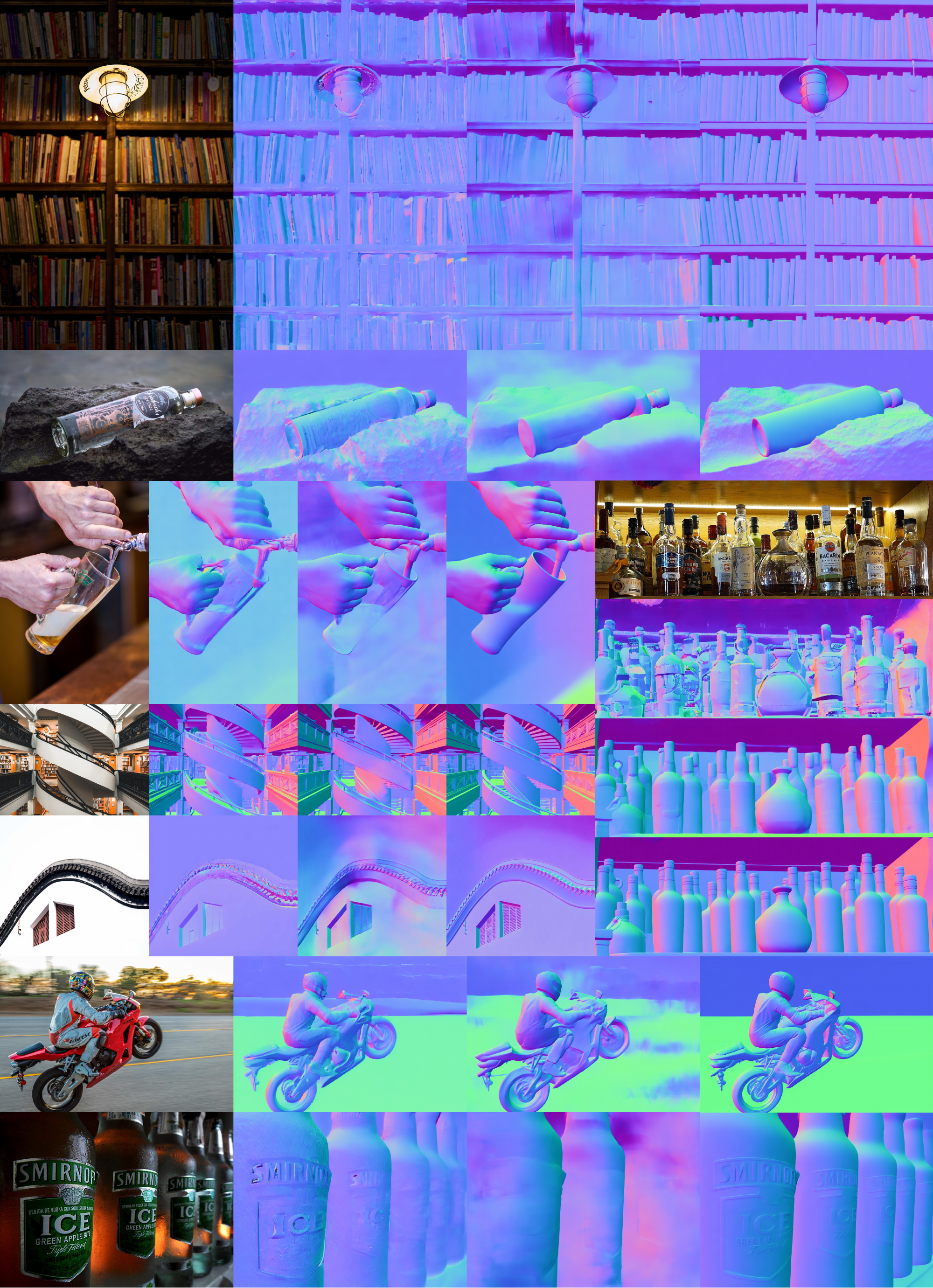}
    \caption{More qualitative results~(Part III). From the left to the right~(the up to the bottom) are results from GeoWizard~\cite{fu2024geowizard}, DSINE\cite{bae2024dsine}, and \modelname respectively.} 
    \label{supp:more-qualitative3}
\end{figure*}

\begin{figure*}[htb] 
    \includegraphics[width=0.9\linewidth]{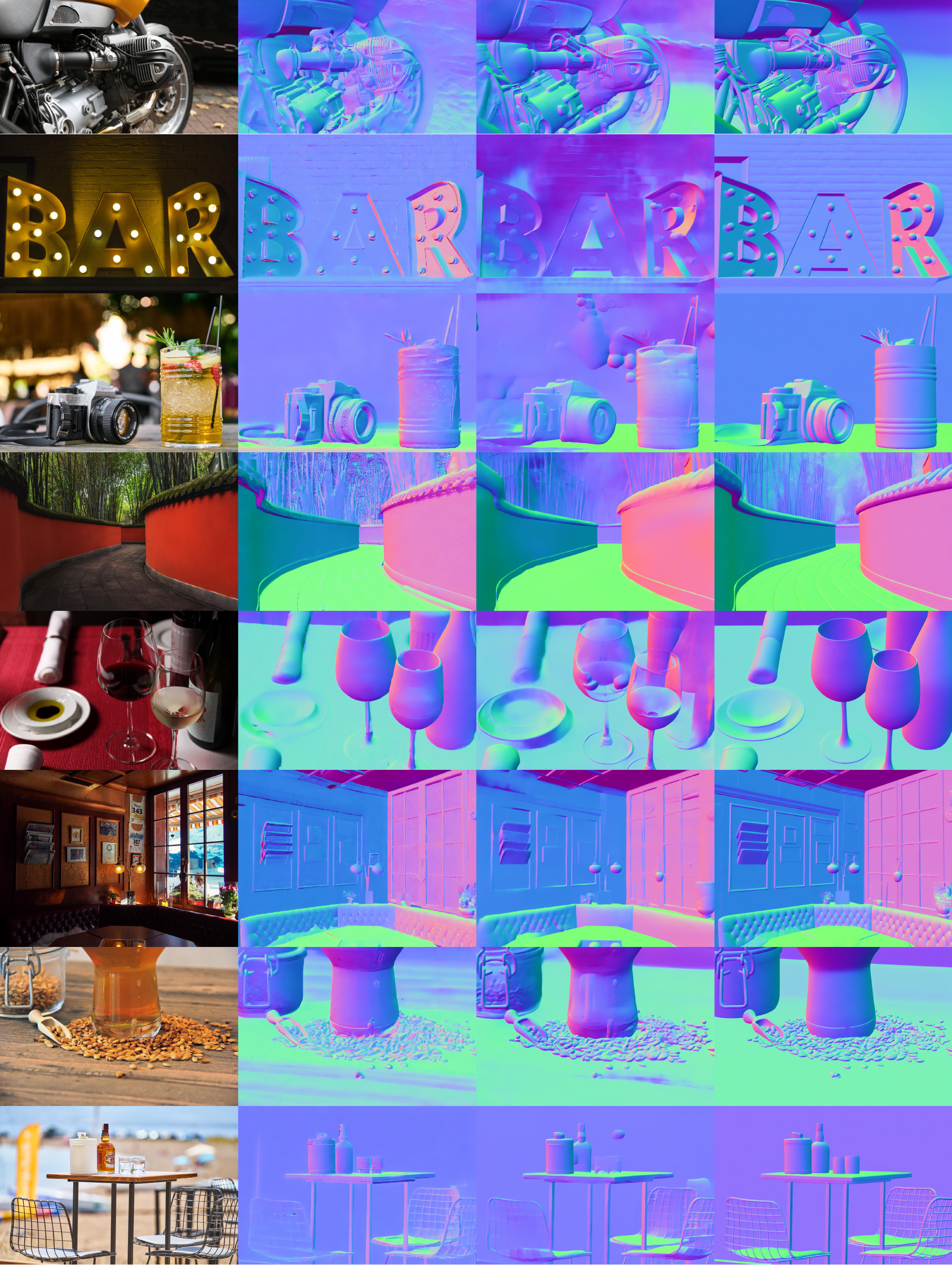}
    \caption{More qualitative results~(Part IV). From the left to the right are results from GeoWizard~\cite{fu2024geowizard}, DSINE~\cite{bae2024dsine}, and \modelname respectively.} 
    \label{supp:more-qualitative4}
\end{figure*}

We present more qualitative comparison results between GeoWizard~\cite{fu2024geowizard}, DSINE~\cite{bae2024dsine}, Marigold~\cite{ke2023marigold}, GenPercept~\cite{xu2024diffusion} and \modelname from \cref{supp:more-qualitative1} to \cref{supp:more-qualitative4}.

\end{subappendices}

\end{document}